\documentclass[10pt,twocolumn,letterpaper]{article}

\usepackage[pagenumbers]{cvpr} 

\usepackage{bm}
\usepackage{algorithm}
\usepackage{algorithmic}
\usepackage{mathrsfs}

\usepackage{colortbl,array,xcolor}

\definecolor{cvprblue}{rgb}{0.21,0.49,0.74}
\usepackage[pagebackref,colorlinks,allcolors=cvprblue]{hyperref}

\def\confName{CVPR}
\def\confYear{2026}
    
\title{Question-guided Visual Compression with Memory Feedback\\for Long-Term Video Understanding}

\author{
    Sosuke Yamao$^{1}$\thanks{Equal contribution} \quad
    Natsuki Miyahara$^{1}$\footnotemark[1] \quad
    Yuankai Qi$^{2}$ \quad
    Shun Takeuchi$^{1,2}$\\
    $^{1}$Fujitsu Research \quad
    $^{2}$Macquarie University
}

\begin{document}
\maketitle
\begin{abstract}
In the context of long-term video understanding with large multimodal models, 
many frameworks have been proposed.
Although transformer-based visual compressors and memory-augmented approaches are 
often used to process long videos, they usually compress each frame 
independently and therefore fail to achieve strong performance on tasks that 
require understanding complete events, such as temporal ordering tasks in MLVU and VNBench. 
This motivates us to rethink the conventional one-way scheme from perception to memory, 
and instead establish a feedback-driven process 
in which past visual contexts stored in the context memory can benefit ongoing perception. 
To this end,
we propose Question-guided Visual Compression with Memory Feedback (QViC-MF),
a framework for long-term video understanding.
At its core is a Question-guided Multimodal Selective Attention (QMSA),
which learns to preserve visual information related to the given question from both the current clip and the past related frames from the memory. The compressor and memory feedback work iteratively for each clip of the entire video.
This simple yet effective design yields large performance gains on long-term video understanding tasks. 
Extensive experiments show
that our method achieves significant improvement over current
state-of-the-art methods by 6.1\% on MLVU test, 8.3\% on LVBench, 18.3\% on VNBench Long, and 3.7\% on VideoMME Long.
%
The code will be released publicly.\footnote{\url{https://github.com/FujitsuResearch/QViC-MF}}
\end{abstract}    
\section{Introduction}
\label{sec:intro}

\begin{figure}[!t] 
  \includegraphics[width=1.0\linewidth]{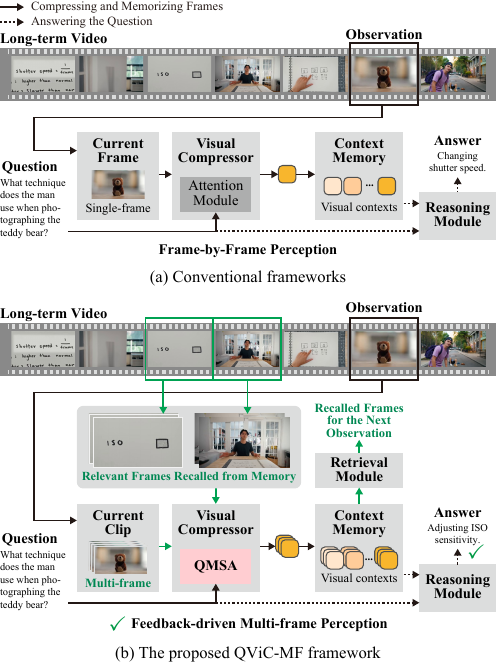}
  \caption{
  (a) Conventional frameworks~\cite{Li2023LLaMAVIDAI, He2024MALMMML, zhang2025llavamini, yamao2024iqvic} 
  compress each frame independently without feedback from memory to perception,
  resulting in limited temporal reasoning and contextual inconsistency.
  (b) The proposed Question-guided Visual Compression with Memory Feedback (QViC-MF) framework 
  performs multi-frame compression guided by both the question and recalled memory,
  enabling feedback-driven perception that preserves temporal event completeness.  
  }
  \label{fig:overview}
\end{figure}

With the rapid advancement of Large Multimodal Models (LMMs), 
their applications have expanded from image-level perception to more complex 
video understanding tasks~\cite{Achiam2023GPT4TR, Reid2024Gemini1U, zhang2024video, bai2025qwen2}.
There is growing interest in utilizing LMMs for long-term video understanding,
which is crucial for real-world applications 
such as analyzing surveillance footage and monitoring industrial processes~\cite{Zhou_2025_CVPR, fu2025video, zhang2025holmes}.
Beyond holistic reasoning, recent benchmarks have introduced increasingly challenging tasks, 
such as the Needle In A Haystack (NIAH) test~\cite{Zhang2024LongCT, zhao2025needle}, 
which demand precise localization of subtle events and reasoning over complex temporal dependencies.

A key challenge is that the limited context window makes it difficult to effectively process entire long videos.
Techniques such as visual token compression~\cite{Li2023LLaMAVIDAI, zhang2025llavamini, shu2025video},
memory augmentation~\cite{Song2023MovieChatFD, He2024MALMMML, Zhang_2025_ICCV, yamao2024iqvic}, and 
frame selection~\cite{tang2025adaptive, yu2025frame} have been proposed to mitigate this issue.
In practice, many recent frameworks combine visual compression with memory augmentation, 
forming a pipeline as illustrated in Figure~\ref{fig:overview}~(a), 
where visual information is first compressed frame by frame with a visual compressor, 
stored in a memory, and then used for output generation in a reasoning module.
This design heavily relies on memory, however, 
which may miss some question-related information due to its limited capacity or improper memory update.

In this work, we propose 
Question-guided Visual Compression with Memory Feedback (QViC-MF),
a long-term video understanding framework that performs multi-frame visual compression guided 
by the question (Figure~\ref{fig:overview}~(b)).
Specifically, we introduce Question-guided Multimodal Selective Attention (QMSA), 
which is the core of our visual compressor. 
QMSA learns to focus on question-related visual information from both the current input clip and the recalled frames from the context memory.
This is achieved via extending the masking mechanism in the self-attention,
where we operate on the attention logits by applying masking, blocking, and guiding strategies.
Regarding the memory feedback, we always retain the Top-$K_\mathrm{r}$ most related frames from all seen clips, 
sorted according to their temporal order. 
This enables our method to have a complete, coherent awareness of the question-related visual content.
Our visual compressor and memory block work in an iterative way until they go through the entire video.
Extensive experiments show a large performance improvement compared to several 
state-of-the-art methods.

The main contributions are summarized as follows:
\begin{itemize}

  \item 
  We propose QViC-MF, an LMM framework for long-term video understanding  
  that builds on QMSA to enable feedback from memory to perception.  
  This design integrates perception and memory through a feedback-driven iterative process,  
  maintaining temporal event completeness and adaptively focusing on informative segments over time.

\item 
  We introduce QMSA, a novel attention mechanism that performs multi-frame and multimodal visual compression  
  conditioned on the question and recalled memory.
  QMSA selectively emphasizes task-relevant cues and forms compact, 
  semantically aligned visual representations.

  \item 
  Extensive experiments on established benchmarks demonstrate that QViC-MF achieves SoTA performance  
  across diverse long-term video understanding tasks.  
  It attains accuracies of 59.4\% on MLVU test~\cite{Zhou_2025_CVPR}, 
  50.3\% on LVBench~\cite{wang2025lvbench}, 63.7\% on VideoMME~\cite{fu2025video}
  and 64.7\% on VNBench~\cite{zhao2025needle}, 
  surpassing existing open-source methods while using fewer visual tokens.  
  These results validate the effectiveness of QViC-MF in achieving both high accuracy and efficiency  
  for long-term video understanding.
\end{itemize}
\section{Related Work}
\label{sec:relatedwork}

\begin{figure*}[!t]
  \centering
  \includegraphics[width=1.0\linewidth]{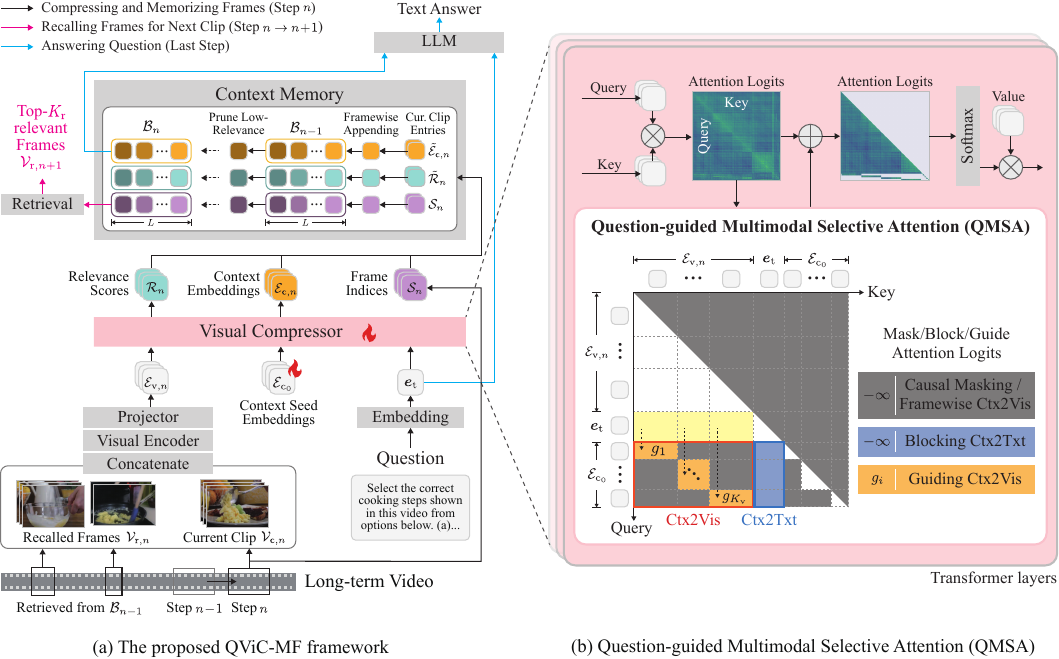}
  \caption{
  (a)~Overview of the proposed Question-guided Visual Compression with Memory Feedback (QViC-MF) framework (Section~\ref{subsec_overall_framework}).
  The video is processed sequentially in clip-wise steps indexed by $n$.
  The visual encoder and projector extract visual embeddings from the current clip $\mathcal{V}_{\mathrm{c},n}$ and recalled frames $\mathcal{V}_{\mathrm{r},n}$ from the context memory $\mathcal{B}_{n-1}$, 
  which are fed into a visual compressor equipped with Question-guided Multimodal Selective Attention (QMSA).
  The compressor transforms the context seed embeddings $\mathcal{E}_{\mathrm{c}_0}$
  into context embeddings $\mathcal{E}_{\mathrm{c},n}$
  by selectively compressing visual features conditioned on the question.
  The context memory $\mathcal{B}_n$ stores the context embeddings with 
  their frame indices $\mathcal{S}_n$ and relevance scores $\mathcal{R}_n$, 
  incrementally appending new entries and pruning low-relevance ones to form the updated memory.
  From this memory, the top-$K_{\mathrm{r}}$ most relevant frame indices are retrieved 
  to form the recalled frames $\mathcal{V}_{\mathrm{r},n+1}$ used for the next clip.
  Finally, the LLM decoder generates the text answer based on the updated memory.
  Trainable components are marked with flame icons.
  (b)~Illustration of QMSA (Section~\ref{subsec_qmsa}).
  QMSA regulates multimodal attention through Mask, Block, and Guide operations,
  enabling frame-wise compression while preserving cross-frame context
  and focusing on question-related features.
  Here, Ctx2Vis and Ctx2Txt denote context-to-visual and context-to-text attention, respectively.
  }
  \label{fig:overview_detail}
\end{figure*}

\noindent\textbf{Long-term Video Understanding}
Understanding long-term videos is challenging due to high computational costs and 
the risk of missing key information. Conventional methods~\cite{Zhang2023VideoLLaMAAI, li2025videochat, Maaz2023VideoChatGPTTD} 
focus on short clips (typically under 1~minute) and do not scale well to longer content. 
To address these issues, recent studies have explored frameworks for efficient long video processing, 
especially for videos exceeding 10~minutes~\cite{zou2024secondshoursreviewingmultimodal}. 
The LLaVA series, including LLaVA-OneVision~\cite{Li2024LLaVAOneVisionEV} and LLaVA-Video~\cite{zhang2024video}, 
extends instruction tuning from images to long-term videos, enhancing multimodal 
understanding with dense temporal coverage.
LongVA~\cite{Zhang2024LongCT} achieves efficient long-term video processing through long context transfer, 
Video-XL~\cite{shu2025video} uses visual summarization tokens to compress hour-long videos while preserving task performance.

\noindent\textbf{Memory-augmented LMMs}
Memory-augmented approaches have also emerged: 
MA-LMM~\cite{He2024MALMMML} processes videos online with a long-term memory bank and 
compression for context retention; MovieChat~\cite{Song2023MovieChatFD} stores short-term and 
long-term memory as transformer tokens.
Flash-VStream~\cite{Zhang_2025_ICCV} enables efficient long-term video understanding and real-time 
response through its core Flash Memory module, integrating Context Synopsis Memory (CSM) and 
Detail Augmentation Memory (DAM) for temporal aggregation.
These approaches perform single-frame compression or use memory in a one-way fashion, 
where visual features are encoded independently and past memory is used only for storage 
and as a reference during decoding, as also observed in IQViC~\cite{yamao2024iqvic}. 
This design limits temporal event completeness and prevents adaptive perception.
The proposed QViC-MF enables both multi-frame compression and feedback-driven visual compression.

\noindent\textbf{Question-aware Visual Encoding}
Question-aware visual encoding,
which adapts visual information to the question posed at inference time,
has been explored in image understanding tasks,
such as Q-Former in InstructBLIP~\cite{Dai2023InstructBLIPTG} 
and QLLaMA in InternVL1.0~\cite{chen2024internvl}.
For long-term video understanding,
MA-LMM~\cite{He2024MALMMML} and LLaMA-VID~\cite{Li2023LLaMAVIDAI} extend these ideas by leveraging 
Q-Former-based query generators to compress visual tokens into lightweight query tokens 
via self-attention and cross-attention layers.
However, these methods perform single-frame compression independently at each time step,
without considering temporal event completeness or referencing past memory during perception.
In contrast, QMSA performs multi-frame compression guided by the question and recalled frames from memory.
\section{Method}
Figure~\ref{fig:overview_detail} shows an overview of the proposed framework.
Our Question-guided Visual Compression with Memory Feedback (QViC-MF) processes long videos sequentially in a clip-by-clip manner,
encoding each clip into visual features with a pre-trained visual encoder,
and projecting them into the visual compressor’s embedding space as visual embeddings via a projector.
The visual compressor equipped with QMSA compresses these visual embeddings into lightweight context embeddings, making them both memory- and question-adaptive.
For subsequent clips, QViC-MF recalls the most relevant frames from the context memory 
based on relevance scores computed from attention weights.
These recalled frames are concatenated with the next clip and fed to the compressor,
enabling feedback from memory to perception and yielding temporally coherent understanding.
Finally, a decoder LLM generates an answer to the question.

In the following, we first describe the overall framework of QViC-MF 
together with its core attention mechanism for visual compression, QMSA, and then explain the training strategy.

\subsection{Question-guided Visual Compression with Memory Feedback}
\label{subsec_overall_framework}

Given the $n$-th clip, QViC-MF first extracts visual embeddings, 
then compresses them into context embeddings using a visual compressor 
built upon QMSA, stores them in the context memory, 
and recalls past frames to guide the compression of the $(n\!+\!1)$-th clip. 
Below, we describe each key component.

\paragraph*{Visual Encoder.}
Given a video $\mathcal{V} = \{\bm{v}_t\}_{t=1}^{T}$, 
where $\bm{v}_t \in \mathbb{R}^{H \times W \times 3}$ represents the $t$-th frame,
QViC-MF divides it into $N$ consecutive clips
$\{\mathcal{V}_{\mathrm{c},i}\}_{i=1}^{N}$, each consisting of $K$ current frames.
For the $n$-th clip $\mathcal{V}_{\mathrm{c},n}$, 
QViC-MF recalls $K_{\mathrm{r}}$ past frames 
$\mathcal{V}_{\mathrm{r},n} \subseteq \{\mathcal{V}_{\mathrm{c},i}\}_{i=1}^{n-1}$ 
based on the previous context memory, where $K_{\mathrm{r}}$ is the number of recalled frames.
The concatenation of $\mathcal{V}_{\mathrm{c},n}$ and $\mathcal{V}_{\mathrm{r},n}$ 
is encoded into visual features by the pre-trained visual encoder $f_{\mathrm{ve}}$
(SigLIP~So400m/14-384px~\cite{zhai2023sigmoid})
and then projected into visual embeddings $\mathcal{E}_{\mathrm{v},n}$
through the projector $f_{\mathrm{proj}}$
(Multi-Layer Perceptron (MLP) projector from LLaVA-Video-7B-Qwen2~\cite{zhang2024video}):
\begin{equation}
  \label{visencoder}
  \mathcal{E}_{\mathrm{v},n}\! = \!
  f_{\mathrm{proj}}\!\left(f_{\mathrm{ve}}\!\left(\operatorname{Concat}\!\left[\mathcal{V}_{\mathrm{c},n}, \mathcal{V}_{\mathrm{r},n}\right]\right)\right)
  \in \mathbb{R}^{K_\mathrm{v} \times P \times D_\mathrm{e}},
\end{equation}
where $K_\mathrm{v} = K + K_\mathrm{r}$, $P$ is the number of patch tokens per frame,
and $D_{\mathrm{e}}$ is the embedding dimension.
For the first clip ($n = 1$), $\mathcal{V}_{\mathrm{r},1}$ is an empty set
since the context memory has not yet been populated, 
and $\mathcal{E}_{\mathrm{v},1}$ therefore consists of $K_\mathrm{v} = K$ frames.

\begin{figure*}[!t]
  \centering
  \includegraphics[width=1.0\textwidth]{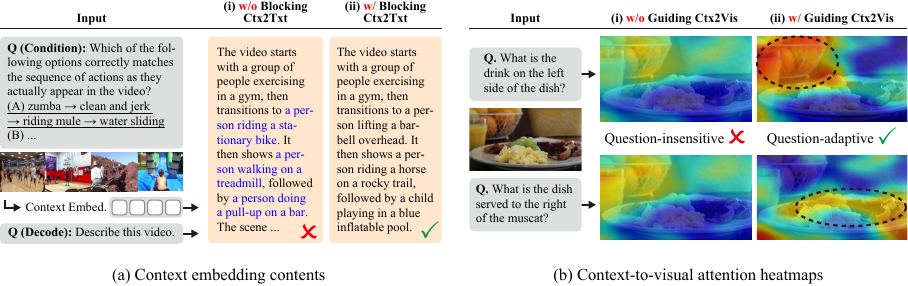}
  \caption{
    (a)
    Without our blocking context-to-text (Ctx2Txt),   
    i.e., when using a naive variant of our visual compressor with standard causal masking in the self-attention module,
    textual information from the question may leak into the context embeddings,
    causing compression hallucination, where the embeddings no longer represent pure visual features.
    (b) Impact of guiding context-to-visual (Ctx2Vis).
    Without guiding Ctx2Vis, the model exhibits question-insensitive attention, failing to adapt its focus to the question.
    Incorporating guiding Ctx2Vis enables question-adaptive attention, allowing the model to attend selectively to regions relevant to the queried content.
  }
  \label{fig:hallucination}
\end{figure*}
\paragraph*{Visual Compressor.}
Unlike existing compressors, which compress each frame in isolation,
our compressor performs question-guided, cross-frame contextual reasoning through self-attention.
The compressor compresses each frame’s $P$ visual patch tokens into 
$C$ context tokens ($C \!\ll\! P$), producing compact per-frame context
embeddings conditioned on the question. 
Our compressor is a transformer-based encoder whose core attention mechanism is QMSA.
QMSA modifies the self-attention logits through three operations: masking, blocking, and guiding.
These operations enable frame-wise compression, suppress compression hallucination, and inject question-adaptive bias.
The detailed formulation of QMSA is given in Section~\ref{subsec_qmsa}.

As shown in Figure~\ref{fig:overview_detail}~(a), 
the input to the visual compressor is the concatenation of the visual embeddings $\mathcal{E}_{\mathrm{v},n}$, 
the question embeddings $\bm{e}_{\mathrm{t}} \in \mathbb{R}^{N_{\mathrm{t}} \times D_{\mathrm{e}}}$,
and the context seed embeddings 
$\mathcal{E}_{\mathrm{c}_0} = \operatorname{Repeat}(\bm{e}_{\mathrm{c}_0}, K_\mathrm{v}) \in \mathbb{R}^{K_\mathrm{v} \times C \times D_{\mathrm{e}}}$,
where $\bm{e}_{\mathrm{c}_0} \in \mathbb{R}^{C \times D_{\mathrm{e}}}$ is a learnable context seed embedding 
repeated $K_\mathrm{v}$ times along the token dimension:
\begin{equation}
  \label{viscompressor_input}
  \mathcal{E}_{\mathrm{in},n} = 
  \operatorname{Concat}\left[\mathcal{E}_{\mathrm{v},n},\ \bm{e}_{\mathrm{t}},\ \mathcal{E}_{\mathrm{c}_0}\right]
  \in \mathbb{R}^{N_\mathrm{enc} \times D_\mathrm{e}},
\end{equation}
where  $N_\mathrm{enc} = K_\mathrm{v}(P+C)+N_{\mathrm{t}}$.
When $\mathcal{E}_{\mathrm{in},n}$ is encoded by the visual compressor $f_{\mathrm{enc}}$,
all tokens are updated while preserving their original ordering:
\begin{equation}
  \label{viscompressor}
  \left[\mathcal{E}'_{\mathrm{v},n},\ \bm{e}'_{\mathrm{t}},\ \mathcal{E}'_{\mathrm{c}_0}\right]
  = f_{\mathrm{enc}}(\mathcal{E}_{\mathrm{in},n}) 
  \in \mathbb{R}^{N_\mathrm{enc} \times D_\mathrm{e}},
\end{equation}
where $\mathcal{E}'_{\mathrm{v},n} \in \mathbb{R}^{K_\mathrm{v} \times P \times D_{\mathrm{e}}}$,
$\bm{e}'_{\mathrm{t}} \in \mathbb{R}^{N_{\mathrm{t}} \times D_{\mathrm{e}}}$,
and $\mathcal{E}'_{\mathrm{c}_0} \in \mathbb{R}^{K_\mathrm{v} \times C \times D_{\mathrm{e}}}$
are the updated embeddings for the visual, text, and context tokens, respectively.
The context embeddings are taken as 
$\mathcal{E}_{\mathrm{c},n} = \{\bm{e}_{\mathrm{c},n,i}\}_{i=1}^{K_\mathrm{v}} = \mathcal{E}'_{\mathrm{c}_0}$,
where each $\bm{e}_{\mathrm{c},n,i} \in \mathbb{R}^{C \times D_{\mathrm{e}}}$ 
corresponds to the $i$-th frame in the concatenated input of $\mathcal{V}_{\mathrm{c},n}$ and $\mathcal{V}_{\mathrm{r},n}$.
%
%

To prepare for storing context embeddings in the context memory and to quantify their relevance to the question, we compute a relevance score for each embedding.
$\mathcal{R}_n = \{r_{n,i}\}_{i=1}^{K_\mathrm{v}} \in \mathbb{R}^{K_\mathrm{v}}$,
where each score reflects cross-frame attention within the visual compressor.
Specifically, each $r_{n,i}$ is obtained by averaging the top-$K_\mathrm{h}$ head-wise
mean attention weights from text tokens to the visual tokens of frame $i$
in the $n$-th clip across layers $L_1$ to $L_2$ of the visual compressor:
\begin{equation}
  r_{n,i} = 
  \operatorname*{Avg}_{l \in [L_1, L_2]}
  \left(
    \operatorname{TopK}\left(\left[\bar{A}_{\mathrm{vt},n,i}^{(l,h)}\right]_{h=1}^H, K_\mathrm{h}\right)
  \right), 
\end{equation}
where $H$ is the number of attention heads,
$\bar{A}_{\mathrm{vt},n,i}^{(l,h)}$ denotes the mean attention weight
from text tokens to the visual tokens of frame $i$ (in clip $n$) at head $h$ of layer $l$,
and $\operatorname{TopK}(\cdot, K_\mathrm{h})$ selects the top-$K_\mathrm{h}$ values.

\paragraph*{Context Memory.}
After compressing the $n$-th clip, we update the context memory in a frame-wise manner. 
Because the context embeddings $\mathcal{E}_{\mathrm{c},n}$ and relevance scores $\mathcal{R}_n$ 
contain entries for both the current clip and the recalled frames, 
we first extract the $K$ entries corresponding to the current clip, denoted as 
$\tilde{\mathcal{E}}_{\mathrm{c},n} = \{ \bm{e}_{\mathrm{c},n,i} \}_{i=1}^{K}$ 
and 
$\tilde{\mathcal{R}}_{n} = \{ r_{n,i} \}_{i=1}^{K}$.
We also record their global frame indices as 
$\mathcal{S}_n = \{ (n-1)K + i \}_{i=1}^{K}$.
The context memory at step $n$ is represented as 
$\mathcal{B}_n = [(\bm{e}_{{\rm m},j}, r_{{\rm m},j}, s_{{\rm m},j})]_{j=1}^{M_n}$, 
where $M_n$ denotes the memory size at step $n$,
and each triplet stores a context embedding, its relevance score, and its global frame index.
The memory capacity is set to $L$. 
For each frame $i = 1, \dots, K$ of the current clip, 
we append the candidate $(\bm{e}_{\mathrm{c},n,i}, r_{n,i}, s_{n,i})$ to the previous memory $\mathcal{B}_{n-1}$.
Whenever the memory size exceeds $L$, 
we prune the entry with the smallest relevance score, 
yielding the updated context memory $\mathcal{B}_n$ at each step.

Finally, to enable feedback from memory to perception, 
we retrieve the top $K_{\mathrm{r}}$ entries with the highest relevance scores from $\mathcal{B}_n$ 
and use their associated frames as the recalled set $\mathcal{V}_{\mathrm{r},n+1}$, 
which is used to process the $(n\!+\!1)$-th clip.

\paragraph*{Decoding.}
Similar to the visual compressor, we employ a transformer-based LLM 
(Qwen2-7B~\cite{yang2024qwen2technicalreport}) as the decoder to generate an answer to the question.
As the final step of the pipeline, 
the text prompt embedding $\bm{e}_{\mathrm{t}}$ is concatenated with 
the context embeddings $\{\bm{e}_{\mathrm{m},j}\}_{j=1}^{M_n}$ 
stored in the context memory $\mathcal{B}_n$ constructed in the previous step.
The resulting sequence is fed to the decoder to generate the answer.

\subsection{Detailed Formulation of Question-guided Multimodal Selective Attention}
\label{subsec_qmsa}

To implement the visual compressor described above, 
we start from a naive self-attention design 
with standard causal masking on long-term video inputs, 
and observe that the text tokens can unintentionally leak information into the context embeddings during compression.
As shown in Figure~\ref{fig:hallucination}~(a)-(i),
this leakage distorts the visual information encoded in the context embeddings, 
causing them to include semantic content that is absent from the corresponding visual features.
We refer to this phenomenon as {compression hallucination}.
Such contamination violates the fundamental requirement that context embeddings remain grounded in the visual evidence,
and constitutes the first challenge addressed in this work.
Another challenge is that 
the context tokens often fail
to attend appropriately to the question when visual and text tokens are fed
jointly into the self-attention module (Figure~\ref{fig:hallucination}~(b)-(i)).
More specifically, when the attention from the context embeddings to the frames
is limited to the context-to-visual scope as shown in Figure~\ref{fig:overview_detail}~(b),
the model struggles to obtain question-adaptive context embeddings.

To address these issues, 
we propose Question-guided Multimodal Selective Attention (QMSA)
to handle multi-frame,
multimodal (text-to-visual) inputs during visual compression.
As shown in Figure~\ref{fig:overview_detail}~(b), 
QMSA regulates the attention pattern inside the visual compressor through three operations: masking, blocking, and guiding. 
Masking enforces causal and framewise constraints,
blocking removes any context-to-text attention that would otherwise cause compression hallucination,
and guiding injects question-dependent biases that steer context embeddings toward question-relevant visual regions.
QMSA controls attention in the visual compressor by applying the masking, blocking, 
and guiding matrices ($\mathbf{M}, \mathbf{B}, \mathbf{G} \in \mathbb{R}^{N_\mathrm{enc} \times N_\mathrm{enc}}$) to 
the attention logits before the softmax operation as follows:
\begin{equation}
\begin{gathered}
  \label{QMSA}
  \mathrm{QMSA}(\mathbf{Q}, \mathbf{K},  \mathbf{V}) = \operatorname{softmax}
  \left( \frac{\mathbf{Q} \mathbf{K}^\top}{\sqrt{D_\mathrm{e}}}\! +\! {\mathbf{M}}\! +\! {\mathbf{B}}\! +\! {\mathbf{G}} \right)\! \mathbf{V},
\end{gathered}
\end{equation}
where $\mathbf{Q}$, $\mathbf{K}$, and $\mathbf{V}$ denote the query, key, and value matrices,
obtained from the input embeddings $\mathcal{E}_{\mathrm{in},n}$ 
through learnable linear projections.

The masking matrix $\mathbf{M}$ extends the general causal masking to handle multi-frame inputs,
allowing the model to capture cross-frame temporal context 
while compressing visual tokens in a framewise manner into their corresponding context tokens.
Each context token attends only to its own visual and context tokens, as follows:
\begin{equation} 
  \mathbf{M}_{i,j} =
  \begin{cases}
  -\infty &
  \begin{aligned}
   &\text{if } i\!<\!j\ \lor\ (i,j)\!\in\mathcal{I}_{\mathrm{c}_k}\!\times\!\{\mathcal{I}_{\mathrm{v}_{\ne k}}, \mathcal{I}_{\mathrm{c}_{<k}}\}, \\
  \end{aligned}
  \\[4pt]
  0 & \text{otherwise.}
  \end{cases}
  \label{eq:mask}
  \end{equation}
\begin{equation}
  \mathcal{I}_x = \{\text{token } i \text{ has label } x \},
  \quad x \in \{ \mathrm{t}, \mathrm{v}_k, \mathrm{c}_k \}.  
\end{equation}
Here, $\mathcal{I}_x$ denotes the set of token indices with label $x$, where the text tokens are labeled as $\mathrm{t}$,
the $k$-th frame visual tokens and context tokens as $\mathrm{v}_{k}$ and $\mathrm{c}_{k}$, respectively. 
$\mathrm{v}_{\ne k}$ and $\mathrm{c}_{< k}$ represent a label set for the visual tokens of frames other than $k$ and context tokens of frames before $k$, respectively.

\begin{table*}[!t]
    \small
    \centering
    \setlength{\tabcolsep}{1.6mm}
    \begin{tabular}{@{}lcc|ccccccc@{}}
        \toprule
         & & {\bf \#Visual} & \multicolumn{2}{|c}{\bf MLVU} & {\bf LVBench} & \multicolumn{2}{c}{{\bf VideoMME} \textit{w/o sub.}}& \multicolumn{2}{c}{{\bf VNBench} \textit{4-try eval.}} \\
        {\bf Method} & {\bf Size (LLM)} & {\bf Tokens} & test & dev & Overall & Overall & Long & Overall & Long \\
        Duration & & & \multicolumn{2}{|c}{3--120 min.} & 30--120 min. & 1--60 min. & 30--60 min. & 10--180 sec. & 60--180 sec. \\
        \midrule
        \multicolumn{4}{l}{\!\!\!\!\textit{Proprietary Models}}  & & & & & \\
        \rowcolor{gray!40} GPT-4o~\cite{Achiam2023GPT4TR} &-&-&54.9&64.6&30.8&71.9&65.3&64.4&56.3\\
        \rowcolor{gray!40} Gemini 1.5 Pro~\cite{Reid2024Gemini1U} &-&-&-&-&33.1&75.0&67.4&66.7&65.1\\
        \midrule
        \multicolumn{4}{l}{\!\!\!\!\textit{Open-source Models (w/o temporal compression)}}  & & & & & \\   
        LLaMA-VID~\cite{Li2023LLaMAVIDAI}&7B (Vicuna-v1.5)&2&17.2&33.2&25.4&25.9&-&10.8&6.4\\
        LLaVA-Mini~\cite{zhang2025llavamini}&7B (Vicuna-v1.5)&1&-&42.8&-& - &-&-&-\\
        LongVA~\cite{Zhang2024LongCT}&7B (Qwen2)&144&41.1&56.3&-&52.6&46.2&-&-\\
        LLaVA-Video~\cite{zhang2024video}&7B (Qwen2)&169&53.3&67.9&41.8&\underline{63.3}&-&62.5&40.4\\
        \midrule
        \multicolumn{4}{l}{\!\!\!\!\textit{Open-source Models (w/ temporal compression)}}  & & & & & \\   
        MovieChat~\cite{Song2023MovieChatFD}&7B (Vicuna-v0)&32&18.0&25.8&21.3&-&-&-&-\\
        MA-LMM~\cite{He2024MALMMML}&7B (Vicuna-v1.1)&32&22.0&36.4&-&-&-&-&-\\
        DynFocus~\cite{han2025dynfocus}&7B (Vicuna-v1.5)&32&-&49.6&31.5&44.1&37.7&-&-\\
        Video-XL~\cite{shu2025video}&7B (Qwen2)&16&45.5&64.9&-&55.5&-&61.6&-\\
        LongVU~\cite{shen2025longvu}&7B (Qwen2)&64&-&65.4&-&60.6&-&-&-\\
        Frame-Voyager~\cite{yu2025frame}&7B (Qwen2)&196&-&65.6&-&57.5&48.9&-&-\\
        Flash-VStream~\cite{Zhang_2025_ICCV}&7B (Qwen2)&128&-&66.3&42.0&61.2&50.3&-&-\\
        \midrule
        \rowcolor{magenta!20} {\bf QViC-MF} (1 fps)&7B (Qwen2)&16&\underline{57.6}&\underline{69.0}& {\bf 50.3} &62.4&\underline{52.6}&\underline{63.0}&\underline{56.9}\\
        \rowcolor{magenta!20} {\bf QViC-MF} (2 fps)&7B (Qwen2)&16&{\bf 59.4}&{\bf 69.6}&\underline{50.2}&{\bf 63.4}&{\bf 54.0}&{\bf 64.7}&{\bf 58.7}\\
        \bottomrule
    \end{tabular}
    \caption{
    Comparison of the proposed QViC-MF framework with SoTA methods on long-term video understanding benchmarks.
    The ``\#Visual Tokens'' column denotes the average number of visual tokens per frame input to the LLM.
    The best and second-best results are highlighted in bold and underlined, respectively.
    }
    \label{tab:benchmark}
    \end{table*}

The blocking matrix $\mathbf{B}$ blocks attention from context tokens to text tokens to prevent compression hallucination.
\begin{equation}
  \mathbf{B}_{i,j} =
  \begin{cases}
  -\infty & \text{if } (i,j) \in \mathcal{I}_{\mathrm{c}_k}\!\times\!\mathcal{I}_{\mathrm{t}}, \\
  0 & \text{otherwise.}
  \end{cases}
\end{equation}
Although this simple blocking effectively suppresses hallucination (Figure~\ref{fig:hallucination}~(a)-(ii)),  
it also reduces the capability to adapt compression to the question (Figure~\ref{fig:hallucination}~(b)-(i)).
To resolve this trade-off and specifically address the second challenge of question-adaptive compression, 
we introduce the guiding matrix $\mathbf{G}$,
which broadcasts guidance from text-to-visual attention to context-to-visual attention as follows:
\begin{equation}
  \mathbf{G}_{i,j} =
  \begin{cases}
  g_j & \text{if } (i,j) \in \mathcal{I}_{\mathrm{c}_k}\!\times\! \mathcal{I}_{{\mathrm{v}}_k}, \\
  0 & \text{otherwise.}
  \end{cases}
\end{equation}
Each element $g_j$ of $\mathbf{G}$ is derived from the attention logits $\mathbf{S}$: 
\begin{equation}
  g_j = \frac{1}{|\mathcal{I}_{\rm{t}}|} \sum_{m \in \mathcal{I}_{\rm{t}}} \mathbf{S}_{m,j}, \quad \mathbf{S} = \frac{\mathbf{Q} \mathbf{K}^\top}{\sqrt{D_{\rm{e}}}}.
\end{equation}

\subsection{Training}
We train two modules: the context seed embedding and the visual compressor.
These modules are optimized to compress video frames into compact spatial representations.
Specifically, we finetune LoRA-adapted LLaVA-Video-7B-Qwen2~\cite{zhang2024video, Hu2021LoRALA} for the visual compressor.
We use an 83K-sample subset randomly drawn from the LLaVA-Video-178K video QA dataset~\cite{zhang2024video},  
ensuring domain balance by sampling 5\% of the original data.
The training is conducted on eight~H200 GPUs. 
For detailed training configurations, please refer to the supplementary material.

\section{Experiments}
We conduct extensive quantitative evaluations of QViC-MF
and compare it with SoTA approaches across multiple long-term video understanding benchmarks. 
In addition, we perform ablation studies to analyze the contribution of each component within QViC-MF.

\subsection{Datasets and Evaluation Metrics}

We evaluate QViC-MF on four complementary benchmarks: 
MLVU~\cite{Zhou_2025_CVPR}, LVBench~\cite{wang2025lvbench}, VideoMME~\cite{fu2025video}, and VNBench~\cite{zhao2025needle}. 
MLVU evaluates diverse long-term video understanding tasks with videos from 3~minutes to over 2~hours, 
containing 2,174 and 502 multiple-choice question-answer pairs (MCQAs) in the dev and test sets, respectively.
Following the official protocol, we report the mean task accuracy (M-Avg).
LVBench serves as a long-term video understanding benchmark with videos up to about two hours (103 videos, 1,549 MCQAs). 
We follow the official scripts and report the overall accuracy across all questions.
VideoMME covers multi-domain video comprehension from short clips to hour-long videos (11~seconds--1~hour), 
consisting of 900 videos and 2,700 MCQAs. 
We evaluate under the video-only setting without subtitles and report the average accuracy across all MCQAs.
VNBench targets the NIAH scenario with 5,400 MCQAs, using a 4-try circular evaluation where a prediction is correct only if all four trials are accurate.

\subsection{Configuration}
We compared QViC-MF with several SoTA methods including LLaVA-Video~\cite{zhang2024video} and Flash-VStream~\cite{Zhang_2025_ICCV}. 
For fair comparison, we reproduced LLaVA-Video results using the authors' publicly released code, 
pretrained models, and official evaluation scripts~\cite{zhang2025lmms}.
For QViC-MF, the number of context tokens per frame $C$ was set to 16, 
and the context memory capacity $L$ was set to 256. 
These values were selected to ensure that the total number of tokens sent to the decoder 
did not exceed the base model's context window size. 
The numbers of clip frames and recall frames $(K, K_r)$ were set to $(32, 32)$. 
As QViC-MF is designed for a streaming-input architecture, 
we first sample video frames at 1 fps or 2 fps. 
The sampled frames are processed sequentially using non-overlapping clips of $K$ frames.
If the total number of sampled frames was less than 64, we instead uniformly sampled 64 frames for input.
All other hyperparameters followed those of the base model.

\subsection{Comparison with SoTA Methods}
As shown in Table~\ref{tab:benchmark},
QViC-MF consistently outperforms all previous open-source methods across multiple benchmarks. 
With 2~fps sampling, QViC-MF achieves the highest overall performance, 
surpassing the previous SoTA by 6.1\% on MLVU test (59.4 vs.\ 53.3), 
by 8.2\% on LVBench (50.2 vs.\ 42.0), by 3.7\% on VideoMME Long (54.0 vs.\ 50.3), and by 18.3\% on VNBench Long (58.7 vs.\ 40.4).

As shown in Table~\ref{tab:benchmark_mlvu_test}, 
a closer examination of the MLVU test results reveals that QViC-MF achieves notable gains in 
Ego Reasoning (ER), Sports QA (SQA), Action Order (AO), and Tutorial QA (TQA), 
indicating an enhanced ability to capture long-term dependencies and temporal relationships.
These improvements remain consistent across different sampling rates, suggesting that 
the proposed question and memory-aware visual compression 
effectively preserves temporal coherence of visual contexts.

Furthermore, QViC-MF attains these results using only 16 context tokens per frame,
which is significantly fewer than LLaVA-Video (169) or Flash-VStream (128),
demonstrating an excellent balance between accuracy and token efficiency, 
which is crucial for scalable long video understanding.

\begin{table}[!t]
    \footnotesize	
    \centering
    \setlength{\tabcolsep}{0.5mm}
    \begin{tabular}{@{}lcccccccccccc@{}}
        \toprule
        {\bf Method} &{\bf Avg} & {\bf TR} & {\bf AR} & {\bf NQA} & {\bf ER} & {\bf PQA} & {\bf SQA} & {\bf AO} & {\bf AC} & {\bf TQA} \\
        \midrule
        \rowcolor{gray!40} GPT-4o &54.9 &83.7 &68.8 &42.9 &47.8 &57.1 &63.6 &46.2 &35.0 &48.7 \\
        \midrule
        LLaMA-VID &17.2 &20.9 &23.1 &21.7 &11.3 &16.0 &16.7 &18.6 &15.0 &11.6 \\
        MovieChat &18.0 &18.7 &10.3 &23.3 &15.1 &16.0 &30.6 &17.1 &15.0 &16.3 \\
        MA-LMM&22.0 &44.0 &23.1 &13.3 &30.2 &14.0 &27.8 &18.6 &13.3 &14.0 \\
        LongVA&41.1 &81.3 &41.0 &46.7 &39.6 &46.0 &44.4 &17.1 &23.3 &30.2 \\
        Video-XL&45.5 &78.0 &28.2 &50.0 &41.5 &46.0 &41.6 &48.6 &{\bf 31.7} &44.2 \\
        LLaVA-Video&53.3 &{\bf 84.6} &\underline{43.6} &\underline{70.0} &\underline{62.3} &\underline{60.0} &38.9 &45.7 &28.3 &46.5 \\
        \midrule
        \rowcolor{magenta!20} {\bf QViC-MF} (1 fps)&\underline{57.6} &\underline{82.4} &{\bf 51.3} &68.3 &{\bf 71.7} &{\bf 62.0} &\underline{52.8} &\underline{50.0} &28.3 &\underline{51.2} \\
        \rowcolor{magenta!20} {\bf QViC-MF} (2 fps)&{\bf 59.4} &\underline{82.4} &\underline{43.6} &{\bf 71.7} &{\bf 71.7}&\underline{60.0} &{\bf 58.3} &{\bf 61.4} &\underline{30.0} &{\bf 55.8} \\
        \bottomrule
    \end{tabular}
    \caption{
    Detailed comparison of QViC-MF with SoTA methods on the MLVU test.
    The best and second-best results are highlighted in bold and underlined, respectively.
    }
    \label{tab:benchmark_mlvu_test}
    \end{table}

\begin{table}[!t]
    \small
    \centering
    \setlength{\tabcolsep}{1.6mm}
    \begin{tabular}{@{}lccccc@{}}
        \toprule
         &  & {\bf MLVU} & \multicolumn{2}{c}{\bf VNBench} \\
        {\bf Settings} & {\bf Input} & test & ALL & Long\\
        \midrule
        Visual Compressor (Vanilla)&64 frm&48.8&61.6&39.3\\
        \rowcolor{orange!20}  + Context Memory&2 fps&42.4&58.2&40.9\\
        \rowcolor{orange!20}  + Memory Feedback&2 fps&52.2&63.7&55.1\\
        \rowcolor{magenta!20} + Framewise Ctx2Vis ($\mathbf{M}$)&2 fps&53.0&61.1&50.7\\
        \rowcolor{magenta!20} + Blocking Ctx2Txt ($\mathbf{B}$)&2 fps&57.9&63.2&56.7\\
        \rowcolor{magenta!20} + Guiding Ctx2Vis ($\mathbf{G}$)&2 fps&{\bf 59.4}&{\bf 64.7}&{\bf 58.7}\\
        \bottomrule
    \end{tabular}
    \caption{
    Ablation study of QViC-MF on MLVU and VNBench.
    Vanilla refers to a naive variant of our visual compressor that uses standard causal masking.
    ``+'' indicates that each component is incrementally added to the Vanilla baseline in the listed order.  
    The best results are highlighted in bold. 
    }
    \label{tab:ablation_design}
    \end{table}

\subsection{Ablation Study}
We conducted several ablation studies to evaluate the contribution of each core component in QViC-MF.

\paragraph{Effect of QViC-MF Design.}
Table~\ref{tab:ablation_design} shows the results of the ablation on the QViC-MF design on MLVU and VNBench.
The Vanilla visual compressor refers to the base model with its input naively 
extended to multiple frames.
Simply adding the context memory, as in existing memory-augmentation methods, 
does not substantially improve performance, whereas applying memory feedback results in a 
clear performance increase in each benchmark.
The QMSA further enhances performance across all configurations.
Each of its variants, namely framewise mask, blocking, and guiding, consistently contributes to accuracy improvements,
showing that explicitly conditioning visual attention on the question and the context memory helps align video representations with task-relevant semantics.
Overall, these results demonstrate that the proposed framework benefits from all key modules in a balanced and mutually supportive manner.


\begin{figure}[!t]
  \centering
  \includegraphics[width=1.0\linewidth]{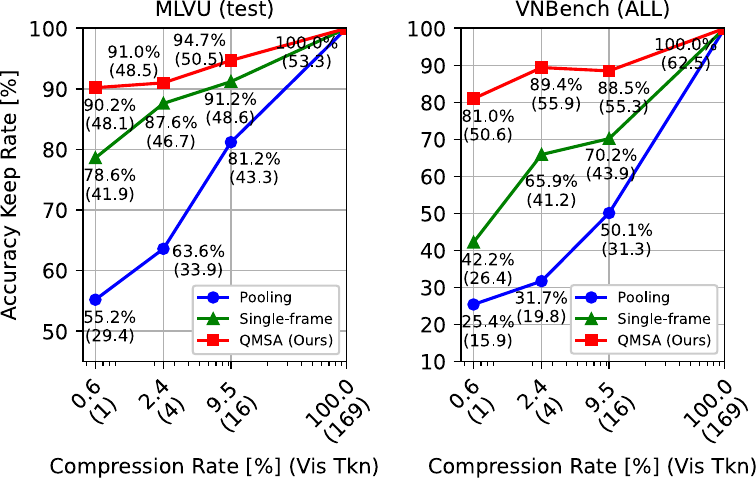}
  \caption{
    Comparison of the proposed visual compressor equipped with QMSA against single-frame and average pooling baseline compressors. 
    ``Single-frame'' denotes frame-by-frame compression without temporal integration. 
    Each point shows the accuracy keep rate (in~\%) with the actual score in parentheses. 
    The horizontal axis represents the compression rate, which is the ratio of visual tokens after compression to those before. 
    To isolate the effect of the visual compressor, context memory and memory feedback are disabled, 
    and all videos are uniformly sampled to 64 frames.
  }
  \label{fig:ablation_compression}
\end{figure}

\paragraph{Comparison of Visual Compressors.}
Figure~\ref{fig:ablation_compression} compares the proposed visual compressor with single-frame and average pooling baseline compressors 
under different compression rates on MLVU and VNBench. 
The baselines show sharp accuracy declines, 
confirming that single-frame or spatially averaged compression fails to preserve semantic structure across frames. 
In contrast, the QMSA-equipped visual compressor effectively compresses video representations 
into compact visual tokens without losing performance. 
Even under high compression, such as using only four or even a single token per frame, 
our visual compressor retains more than 90\% and 80\% of its full-token accuracy on MLVU and VNBench, respectively, 
whereas the baselines suffer severe performance drops. 
This indicates that the visual compressor preserves task-relevant visual contexts while discarding redundant information.

\begin{figure}[!t]
    \centering
    \includegraphics[width=1.0\linewidth]{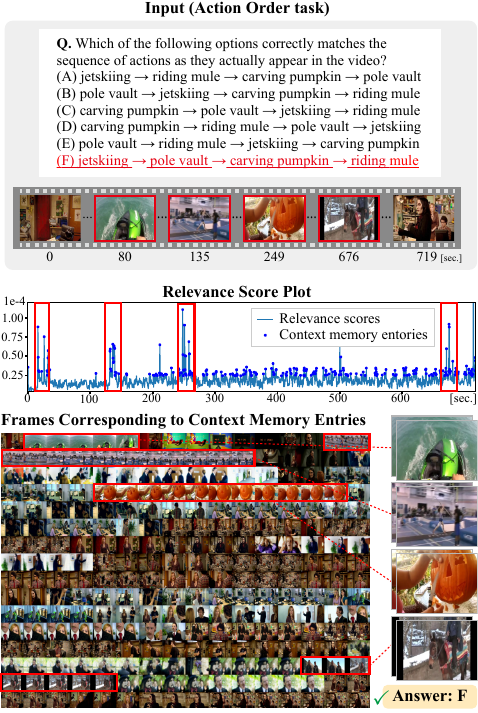}
    \caption{
    Case study example from MLVU, illustrating how QViC-MF utilizes context memory to answer long-term reasoning questions. 
    The example corresponds to an Action Order (AO) task question, 
    where the model predicts the correct sequence of events over a long-term video. 
    Input frames corresponding to stored context memory entries are shown, and answer-critical frames are highlighted in red.
    The relevance score plot indicates the relevance score of each context memory entry over time.
    }
    \label{fig:result_example}
    \vspace{-1.5mm} 
  \end{figure}

\subsection{Case Study}
Figure~\ref{fig:result_example} shows how QViC-MF utilizes its context memory to perform long-term reasoning.
The example corresponds to an Action Order question from MLVU, where the model must determine the correct temporal sequence of four actions occurring across a 12-minute video.
As illustrated in Figure~\ref{fig:result_example}, 
the answer-critical frames required to solve the task are highlighted in red,
corresponding to key events distributed across the entire video timeline.
The relevance score plot shows that QViC-MF assigns high relevance scores to these temporally sparse but semantically important moments.
Consequently, these key frames are retained in the context memory throughout the streaming input.
Because the context memory preserves these answer-critical frames, the decoder can refer to them when reasoning about the global event order, enabling QViC-MF to correctly predict the action sequence.
Overall, this qualitative example demonstrates that QViC-MF effectively identifies and preserves task-relevant visual contexts while discarding redundant frames, allowing efficient and interpretable long-term video understanding.
Details of the relevance score analysis are provided in the supplementary material.

\vspace{-1.5mm} 
\section{Conclusion}
In this work, we propose QViC-MF, a framework for long-term video understanding. QViC-MF performs multi-frame visual compression guided 
by the question and benefits from memory feedback.
%
%
This iterative feedback-driven design bridges perception and memory, 
maintaining temporal event completeness and enhancing reasoning across long-term videos. 
Extensive evaluations on MLVU, LVBench, VideoMME, and VNBench 
demonstrate that QViC-MF achieves state-of-the-art accuracy,
while requiring fewer visual tokens per frame. 
Future work will explore extensions to online and streaming scenarios,
with a particular focus on continuous visual memory as a foundation
for further advancing multimodal agentic AI systems.

{
    \small
    \bibliographystyle{ieeenat_fullname}
    \bibliography{main}
}

\maketitlesupplementary
\appendix

This supplementary material provides: 
(i)~detailed training setup and hyper-parameters 
of the Question-guided Visual Compression with Memory Feedback (QViC-MF) framework, 
as well as the inference configuration (Section~\ref{sec:training_details}); 
(ii)~a description of the masking strategies used in the visual compressor and their relation to 
Question-guided Multimodal Selective Attention (QMSA) (Section~\ref{sec:masking_visual_compressor});
(iii)~an analysis of the hyper-parameters for relevance score computation (Section~\ref{sec:relevance_score_computation});
(iv)~full benchmark results on MLVU-dev, LVBench, VideoMME, and VNBench (Section~\ref{sec:benchmark_details});
(v)~additional ablation studies on feedback frames, relevance computation, context tokens, and memory capacity (Section~\ref{sec:ablation_studies});
(vi)~a computational complexity analysis of QViC-MF (Section~\ref{sec:complexity});
(vii)~a report on inference time and VRAM usage (Section~\ref{sec:inference_time_vram});
(viii)~an analysis of the reliability of relevance-based memory retrieval 
using needle hit-rate evaluation on VNBench Long 
(Section~\ref{sec:memory_stability});
and (ix) a discussion of task-dependent behavior, limitations, and failure cases (Section~\ref{sec:limitation}).

\section{Training Details}
\label{sec:training_details}

\paragraph{Dataset.}
We use an 83K-sample subset obtained by randomly sampling 5\% of the 
LLaVA-Video-178K dataset~\cite{zhang2024video}, 
which contains a wide range of instruction-following tasks such as captioning, open-ended VQA, and multiple-choice VQA. 
We adopt this subset to ensure a manageable computational footprint while maintaining the 
task diversity of the original dataset. 

\paragraph{Protocols.}
The training hyper-parameters are summarized in Table~\ref{tab:hyperparameter_train}.
We train two modules: the visual compressor and the context seed embeddings, 
while keeping all other components frozen. 
LoRA~\cite{Hu2021LoRALA} is applied to the base model of the visual compressor (LLaVA-Video-7B-Qwen2~\cite{zhang2024video}).
The decoder LLM (Qwen2-7B~\cite{yang2024qwen2technicalreport}), which remains frozen during training,
is also based on the LLaVA-Video-7B-Qwen2 architecture.
During training, the model operates in a one-way compression setting without applying feedback from the context memory. 
Accordingly, we set the number of clip frames to $K = 64$ and the number of recalled frames to $K_{\mathrm{r}} = 0$, 
so that the visual compressor receives $K_{\mathrm{v}} = K + K_{\mathrm{r}} = 64$ frames, 
matching the input specification of the base model (LLaVA-Video-7B-Qwen2). 

For each sample from LLaVA-Video-178K, we uniformly sample $K$ frames from the video and feed their visual embeddings,
together with the question embeddings, into the visual compressor to obtain $K$ context embeddings.
To avoid overfitting the decoder LLM to a fixed and relatively small memory size, 
and to address the fact that the $K$ sampled frames do not fill the entire memory when $L > K$ (with $L = 256$), 
we randomize the effective memory length during training.  
Specifically, we interpolate the $K$ context embeddings using nearest-neighbor interpolation to a sequence length uniformly 
sampled between $K$ and $L$, and use this sequence as the content of the context memory.
Finally, the context memory and the question are fed to the decoder to generate an answer, and a standard cross-entropy 
loss with respect to the ground-truth answer is computed. Minimizing this loss updates the trainable context seed embeddings 
and the parameters of the visual compressor.

The inference-time configuration, including the hyper-parameters for relevance score computation, 
is summarized in Table~\ref{tab:hyperparameter_inference}, and the overall inference 
pipeline follows the procedure described in Section~3 of the main paper.

\begin{table}[!t]
    \small
    \centering
    \begin{tabular}{@{}lcc@{}}
      \toprule
      \textbf{Setting} &  &  \\
      \midrule
      Batch size                    & 1  \\
      Gradient accumulation steps   & 4  \\
      Learning rate                 & 1e-4 \\
      Learning scheduler            & Cosine decay \\
      Warm-up ratio                 & 0.03 \\
      Weight decay                  & 0 \\
      Number of epochs              & 1 \\
      Optimizer                     & AdamW \\
      DeepSpeed stage               & 3  \\
      LoRA rank                     & 64 \\
      LoRA alpha                    & 16 \\
      LoRA dropout                  & 0.05 \\
      Context tokens $C$            & 16   \\
      Context memory capacity $L$   & 256 \\
      Clip frames $K$               & 64 \\
      Recalled frames $K_{\mathrm{r}}$ & 0 \\
      Visual encoder                & Frozen \\
      Context seed embedding        & Trainable \\
      Visual compressor             & Trainable \\
      Decoder LLM                   & Frozen  \\
      \bottomrule
    \end{tabular}
    \caption{
    Training settings of our QViC-MF framework.
    During training, the model operates in a one-way compression setting without applying feedback 
    from the context memory, and thus the number of recalled frames is fixed to $K_{\mathrm{r}} = 0$.
    }
    \label{tab:hyperparameter_train}
  \end{table}

\begin{table}[!t]
    \small
    \centering
    \begin{tabular}{@{}lcc@{}}
      \toprule
      \textbf{Setting} &  &  \\
      \midrule
      Context tokens $C$            & 16   \\
      Context memory capacity $L$   & 256 \\
      Clip frames $K$               & 32 \\
      Recalled frames $K_{\mathrm{r}}$ & 32 \\
      Top heads $K_\mathrm{h}$ for computing $r_{n,i}$     & 5 \\
      Layer range $[L_1, L_2]$ for computing $r_{n,i}$ & [17, 20] \\
      \bottomrule
    \end{tabular}
    \caption{
    Inference settings of QViC-MF, including the hyper-parameters
    for relevance score computation. Parameters shared with
    training settings are repeated for clarity.
    }
    \label{tab:hyperparameter_inference}
  \end{table}

\begin{figure*}[!t]
  \centering
  \includegraphics[width=1.0\linewidth]{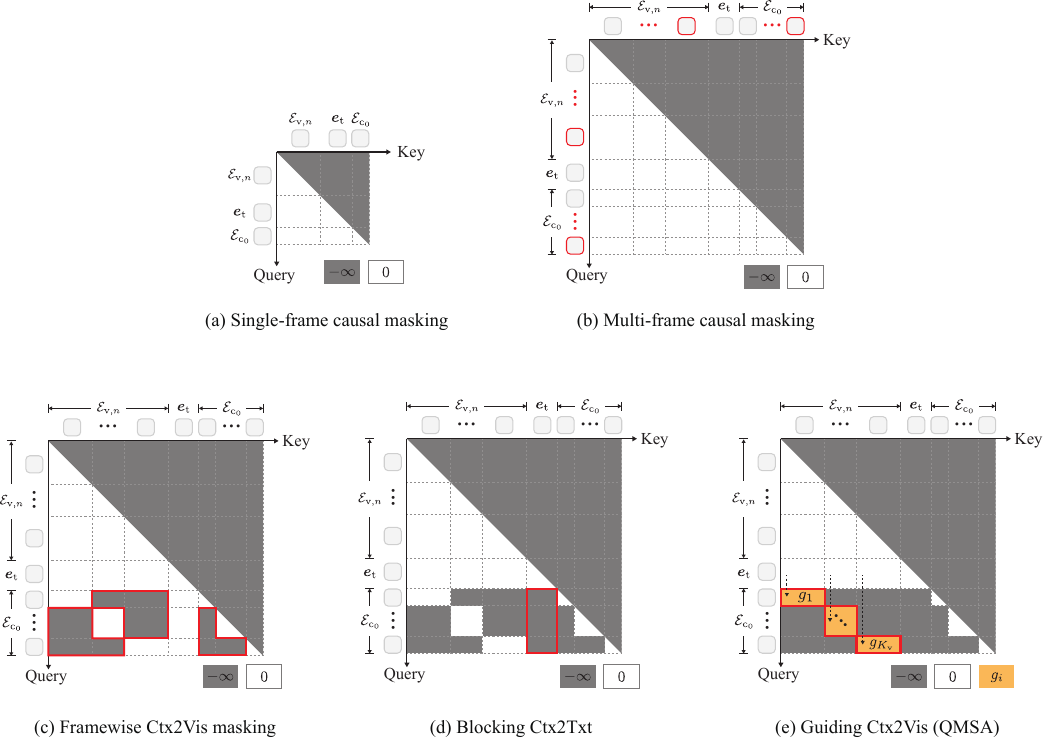}
  \caption{
    Variations of masking strategies used in the visual compressor.
    (a)~A simple baseline that performs conventional frame-to-frame compression using single-frame causal masking.
    This corresponds to the ``Single-frame’’ setting in Figure~4 of the main paper.
    (b)~A naive extension of (a) to multi-frame compression,
    corresponding to the ``Vanilla’’ setting in Table~3 of the main paper.
    (c)~A variant of (b) that applies framewise context-to-visual masking, 
    matching the masking matrix $\mathbf{M}$ in QMSA.
    (d)~A variant of (c) that additionally blocks context-to-text attention,
    corresponding to applying both $\mathbf{M}$ and the blocking matrix $\mathbf{B}$ in QMSA.
    (e)~The complete QMSA configuration, which further introduces guiding for
    context-to-visual attention. This corresponds to applying all three matrices:
    $\mathbf{M}$, $\mathbf{B}$, and $\mathbf{G}$.
    As in standard attention masking, $-\infty$ blocks token-pair attention (ignored after softmax), 
    $0$ passes the logits unchanged, and the guiding values $g_i$ act as bias terms added to the attention logits.
  }
  \label{fig:masking_variants}
\end{figure*}

\section{Masking Strategies in the Visual Compressor}
\label{sec:masking_visual_compressor}

Figure~\ref{fig:masking_variants} illustrates several masking strategies used in the proposed visual compressor.
These masking schemes determine how each token attends to past or future visual, text, and context tokens across frames.
The strategies play a crucial role not only in controlling the temporal receptive field 
but also in preventing information leakage that could lead to compression hallucination.
Moreover, these strategies provide the structural foundations required for question-adaptive visual compression.
In particular, our proposed Question-guided Multimodal Selective Attention (QMSA) builds on these masking 
components to selectively preserve information relevant to the given question 
while avoiding undesirable cross-frame or cross-modal interactions.

\paragraph{Single-frame Causal Masking (Figure~\ref{fig:masking_variants}~(a)).}
The single-frame causal masking corresponds to conventional frame-to-frame attention.
Each token can attend only to tokens within the same frame and only to those occurring earlier in temporal order.
While this avoids any cross-frame information mixing, it also prevents the model from capturing temporal dependencies across frames.
As a result, the visual compressor cannot exploit temporal continuity or motion cues,
limiting its ability to perform temporally informed compression.
In addition, because this configuration disallows any interaction across frames, 
it cannot support memory-feedback-based visual compression as used in QViC-MF.
This configuration serves as a baseline and matches the ``Single-frame’’ setting shown in Figure~4 of the main paper.

\paragraph{Multi-frame Causal Masking (Figure~\ref{fig:masking_variants}~(b)).}
A straightforward extension is the multi-frame causal masking, 
where each token may attend to all tokens from past frames while future frames remain masked.
Although this enables temporal modeling and, in principle, allows the visual compressor to accept 
memory-feedback inputs across frames, 
it causes the context embeddings to become an entangled representation of the entire input clip rather than frame-local representations.
As a result, a single context embedding no longer corresponds to the visual content of an individual frame, 
causing frame-wise addition or removal to produce a context memory whose frame-level semantics collapse.
This configuration corresponds to the ``Vanilla’’ setting reported in Table~3 of the main paper.

\paragraph{Framewise Context-to-Visual Masking (Figure~\ref{fig:masking_variants}~(c)).}
The framewise context-to-visual masking matrix~$\mathbf{M}$ addresses the limitations of single-frame and multi-frame causal masking.
This mask enables the visual compressor to capture temporal relationships during compression 
while keeping each context embedding as an independent frame-local representation.
However, as shown in Figure~3~(a) of the main paper, text information may still leak into the context tokens, 
distorting the visual content of the context embeddings and causing compression hallucination.

\paragraph{Blocking Context-to-Text Masking (Figure~\ref{fig:masking_variants}~(d)).}
The blocking matrix~$\mathbf{B}$ prevents context-to-text attention and mitigates compression hallucination.
However, as shown in Figure~3~(b) of the main paper, removing text-context interaction hinders question-adaptive compression 
and reduces the model’s ability to preserve information relevant to the given question.

\paragraph{Guiding Context-to-Visual Masking (QMSA; Figure~\ref{fig:masking_variants}~(e)).}
The full QMSA configuration is obtained by introducing a guiding matrix~$\mathbf{G}$ in addition to $\mathbf{M}$ and $\mathbf{B}$.
The matrix~$\mathbf{G}$ enables controlled text-to-visual information flow into the context tokens, 
allowing the visual compressor to perform question-adaptive compression.
With this design, QMSA resolves the limitations of the previous masking strategies and supports context-aware visual compression 
while maintaining temporal causality and preventing undesired cross-modal interference.

\begin{figure*}[!t]
  \centering
  \includegraphics[width=1.0\linewidth]{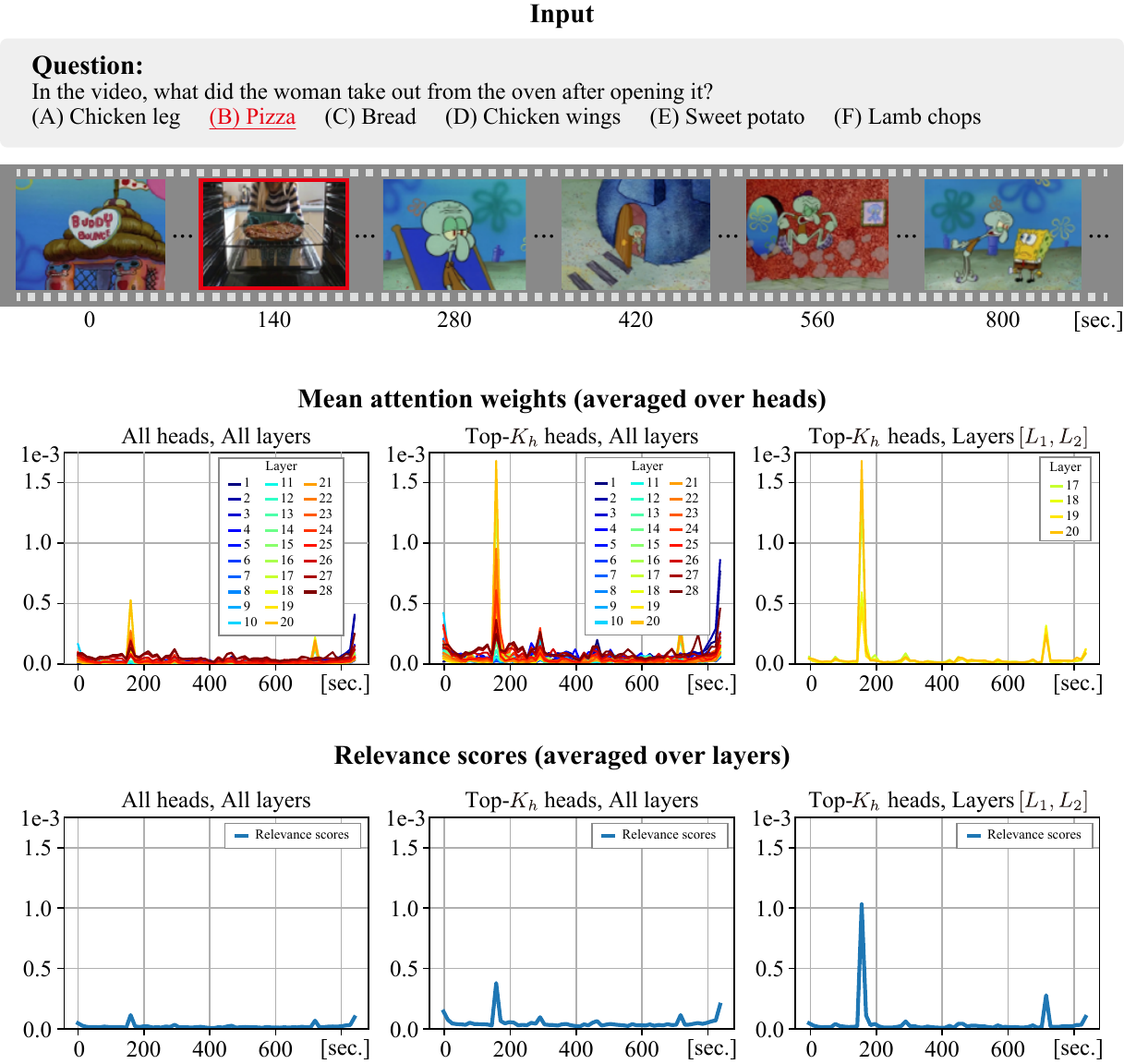}
  \caption{
    Visualization of attention patterns and relevance scores 
    for a sample from the Needle QA (NQA) task in MLVU.
    The frames relevant to the question are highlighted with red boxes (around 140 seconds).
    The middle row shows the mean text-to-visual attention weights: 
    for each condition (all heads vs.\ top-$K_\mathrm{h}\!=\!5$ heads, and all layers vs.\ 
    layers $L_1\!=\!17$ to $L_2\!=\!20$),
    we plot the head-averaged attention weights for each layer,
    with different colors indicating different layers.
    The bottom row shows the corresponding relevance scores obtained by 
    averaging these layer-wise mean attention weights across layers.
    These results show that appropriately selecting heads and layers 
    leads to more discriminative relevance scores aligned with the question.
    For visualization, we assign each visual token to the frame 
    it originates from based on its fixed sequence position 
    and aggregate the text-to-visual attention over the tokens of each frame. 
    This provides an approximate yet practical frame-level attribution even in deep layers.
  }
  \label{fig:attn_weights_for_relevance}
\end{figure*}

\begin{table*}[!t]
  \small
  \centering
  \begin{tabular}{@{}lcc|cccccccc@{}}
    \toprule
      {\bf Model} & {\bf Size (LLM)} & {\bf Input} & {\bf M-Avg} & {\bf TR} & {\bf AR} & {\bf NQA} & {\bf ER} & {\bf PQA} & {\bf AO} & {\bf AC}\\
    \midrule
      \rowcolor{gray!40} GPT-4o~\cite{Achiam2023GPT4TR} &-&0.5 fps&64.6 &-&-&-&-&-&-&-\\
      MovieChat~\cite{Song2023MovieChatFD}&7B (Vicuna-v0)&2048 frm&25.8 &29.5 &25.0 &24.2 &24.7 &25.8 &28.6 &22.8 \\
      LLaMA-VID~\cite{Li2023LLaMAVIDAI}&7B (Vicuna-v1.5)&1 fps&33.2 &50.8 &34.5 &30.1 &32.7 &32.5 &23.9 &27.8 \\
      MA-LMM~\cite{He2024MALMMML}&7B (Vicuna-v1.1)&1000 frm&36.4 &51.9 &35.5 &43.1 &38.9 &35.8 &25.1 &24.3 \\
      LLaVA-Mini~\cite{zhang2025llavamini}&7B (Vicuna-v1.5)&1 fps&42.8 &76.0 &50.0 &44.5 &37.5 &49.0 &24.3 &18.4 \\
      DynFocus~\cite{han2025dynfocus}&7B (Vicuna-v1.5)&32 frm&49.6 &76.2 &60.9 &55.5 &41.5 &54.0 &26.8 & \underline{32.8} \\
      LongVA~\cite{Zhang2024LongCT} &7B (Qwen2)&256 frm&56.3 &-&-&-&-&-&-&-\\
      Video-XL~\cite{shu2025video}&7B (Qwen2)&256 frm&64.9 &-&-&-&-&-&-&-\\
      LongVU~\cite{shen2025longvu}&7B (Qwen2)&1 fps&65.4 & {\bf 87.5} & {\bf 76.0} &76.3 &59.4 &71.6 &58.3 &29.0 \\
      Frame-Voyager~\cite{yu2025frame}&7B (Qwen2)&8 frm&65.6 &-&-&-&-&-&-&-\\
      Flash-VStream~\cite{Zhang_2025_ICCV}&7B (Qwen2)&1 fps&66.3 &-&-&-&-&-&-&-\\
      LLaVA-Video~\cite{zhang2024video}&7B (Qwen2)&64 frm&67.9 & \underline{85.9} & 67.5 &80.3 &67.1 &75.3 &58.3 & {\bf 40.8} \\
      \rowcolor{magenta!20}{\bf QViC-MF (Ours)}&7B (Qwen2)&1 fps& \underline{69.0} & 84.0 &66.5 & \underline{84.2} & \underline{71.3} & \underline{76.4} & \underline{68.3} & 32.0 \\
      \rowcolor{magenta!20}{\bf QViC-MF (Ours)}&7B (Qwen2)&2 fps& {\bf 69.6} &84.4 & \underline{68.5} & {\bf 84.8} & {\bf 72.7} & {\bf 76.8} & {\bf 69.5} &30.1 \\
    \midrule
  \end{tabular}
  \caption{
    Detailed comparison of the proposed QViC-MF framework with SoTA methods on the MLVU-dev.
    The best and second best results are highlighted in bold and underlined, respectively.
    }
    \label{tab:mlvudev}
\end{table*}

\begin{table*}[!t]
  \small
  \centering
  \begin{tabular}{@{}lcc|ccccccc@{}}
    \toprule
      {\bf Model} & {\bf Size (LLM)} & {\bf Input} & {\bf Overall} & {\bf KIR} & {\bf EU} & {\bf ER} & {\bf TG} & {\bf Rea} & {\bf Sum}\\
    \midrule
    \rowcolor{gray!40} GPT-4o~\cite{Achiam2023GPT4TR} &-&348 frm&30.8 &34.5 &27.4 &33.0 &25.0 &27.5 &24.1 \\
    \rowcolor{gray!40} Gemini 1.5 Pro~\cite{Reid2024Gemini1U} &-&3600 frm&33.1 &39.3 &30.9 &32.1 &31.8 &27.0 &32.8 \\
    MovieChat~\cite{Song2023MovieChatFD}&7B (Vicuna-v0)&2048 frm&22.5 &25.9 &23.1 &21.3 &22.3 &24.0 &17.2 \\
    LLaMA-VID~\cite{Li2023LLaMAVIDAI}&7B (Vicuna-v1.5)&1 fps&23.9 &23.4 &21.7 &25.4 &26.4 &26.5 &17.2 \\
    DynFocus~\cite{han2025dynfocus}&7B (Vicuna-v1.5)&200 frm&31.5 &31.3 &32.6 &30.1 &25.5 &33.3 &30.5 \\
    LLaVA-Video~\cite{zhang2024video}&7B (Qwen2)&64 frm&41.8 &41.6 & \underline{39.3} &42.3 &33.2 &47.8 & {\bf 32.8} \\
    Flash-VStream~\cite{Zhang_2025_ICCV}&7B (Qwen2)&1 fps&42.0 &-&-&-&-&-&-\\
    \rowcolor{magenta!20}{\bf QViC-MF (Ours)}&7B (Qwen2)&1 fps& {\bf 50.3} & {\bf 54.6} & {\bf 47.5} & \underline{52.3} & {\bf 41.4} & \underline{50.8} & \underline{31.0} \\
    \rowcolor{magenta!20}{\bf QViC-MF (Ours)}&7B (Qwen2)&2 fps& \underline{50.2} & \underline{54.3} & {\bf 47.5} & {\bf 52.4} & \underline{40.0} & {\bf 51.2} & 29.3 \\
    \midrule
  \end{tabular}
  \caption{
    Detailed comparison of the proposed QViC-MF framework with SoTA methods on the LVBench.
    The best and second best results are highlighted in bold and underlined, respectively.
    }
    \label{tab:lvbench}
\end{table*}

\section{Effects of Hyper-parameters in Relevance Score Computation}
\label{sec:relevance_score_computation}

Figure~\ref{fig:attn_weights_for_relevance} illustrates the effects of two key hyper-parameters in our 
relevance score computation: the number of top attention heads $K_\mathrm{h}$ 
and the layer range $[L_1, L_2]$ from which the text-to-visual attention weights are extracted.  
We adopt the inference-time configuration (Table~\ref{tab:hyperparameter_inference}), 
namely $K_\mathrm{h} = 5$ and $[L_1, L_2] = [17, 20]$.
We use a sample from the Needle QA (NQA) task in MLVU. 
The frames relevant to the question (around 140 seconds) are highlighted in red,
while all other frames are unrelated to the query.

The middle row of Figure~\ref{fig:attn_weights_for_relevance} shows the mean text-to-visual attention weights. 
For each condition (all heads vs.\ top-$K_\mathrm{h}$ heads and all layers vs.\ 
layers $L_1$--$L_2$), we plot the head-averaged attention weights for each layer,
with different colors corresponding to different layers.
Focusing on the heads that respond strongly to the question leads to 
more discriminative attention patterns across layers. 
In addition, we observe that text-visual interactions become particularly pronounced in the middle-to-late layers, 
suggesting that this layer range contributes more strongly to identifying question-relevant frames.
The bottom row shows the relevance scores obtained by averaging these layer-wise mean attention weights across layers.
Using all heads and all layers yields noisy and weakly discriminative signals, 
whereas selecting $K_\mathrm{h}$ and layers $L_1$--$L_2$ produces sharp peaks precisely at the 
question-relevant frames, resulting in much clearer relevance scores.

\begin{table*}[!t]
  \small
  \centering
  \begin{tabular}{@{}lcc|cccc@{}}
    \toprule
      {\bf Model} & {\bf Size (LLM)} & {\bf Input} & {\bf Overall} & {\bf Short} & {\bf Medium} & {\bf Long}\\
    \midrule
      \rowcolor{gray!40} GPT-4o~\cite{Achiam2023GPT4TR} &-&384 frm&71.9 &80.0 &70.3 &65.3 \\
      \rowcolor{gray!40} Gemini 1.5 Pro~\cite{Reid2024Gemini1U} &-&1 fps&70.3 &78.8 &68.8 &61.1 \\
      DynFocus~\cite{han2025dynfocus}&7B (Vicuna-v1.5)&16 frm&44.1 &50.9 &43.7 &37.3 \\
      LongVA~\cite{Zhang2024LongCT} &7B (Qwen2)&128 frm&52.6 &61.1 &50.4 &46.2 \\
      Video-XL~\cite{shu2025video}&7B (Qwen2)&128 frm&55.5 &64.0 &53.2 &49.2 \\
      Frame-Voyager~\cite{yu2025frame}&7B (Qwen2)&8 frm&57.5 &67.3 &56.3 &48.9 \\
      Flash-VStream~\cite{Zhang_2025_ICCV}&7B (Qwen2)&1 fps&61.2 &72.0 &\underline{61.1} &50.3 \\
      LLaVA-Video~\cite{zhang2024video}&7B (Qwen2)&64 frm& \underline{63.3} &-&-&-\\
      \rowcolor{magenta!20}{\bf QViC-MF (Ours)}&7B (Qwen2)&1 fps&62.4 & \underline{74.3} &60.4 &\underline{52.6} \\
      \rowcolor{magenta!20}{\bf QViC-MF (Ours)}&7B (Qwen2)&2 fps& {\bf 63.4} & {\bf 74.6}  & {\bf 61.6} & {\bf 54.0} \\
    \midrule
  \end{tabular}
  \caption{
    Detailed comparison of the proposed QViC-MF framework with SoTA methods on the VideoMME without subtitle setting. 
    The best and second best results are highlighted in bold and underlined, respectively.
    }
    \label{tab:videomme}
\end{table*}

\begin{table*}[!t]
  \footnotesize	
  \centering
  \setlength{\tabcolsep}{1.0mm}
  \begin{tabular}{@{}lcc|c|cccc|cccc|cccc@{}}
      \toprule
      & & & {\bf Overall} & \multicolumn{4}{c}{\bf Retrieval} & \multicolumn{4}{c}{\bf Ordering} & \multicolumn{4}{c}{\bf Counting} \\
      {\bf Method} & {\bf Size (LLM)} & {\bf Input} & & Edit & Insert-1 & Insert-2 & Avg & Edit & Insert-1 & Insert-2 & Avg & Edit-1 & Edit-2 & Insert & Avg \\
      \midrule
      \multicolumn{4}{l}{\!\textit{ALL (10-180 sec.)}}  && \\
      \rowcolor{gray!40} GPT-4o~\cite{Achiam2023GPT4TR} &-&1 fps &64.4 &100.0 &98.0 &87.3 &95.3 &88.4 &86.6 &45.2 &73.4 &36.8 &0.0 &36.1 &24.5 \\
      \rowcolor{gray!40} Gemini 1.5 Pro~\cite{Reid2024Gemini1U} &-&1 fps&66.7 &100.0 &96.0 &76.0 &90.7 &90.7 &95.3 &32.7 &72.9 &60.7 &7.3 &42.0 &36.7 \\
        LLaMA-VID~\cite{Li2023LLaMAVIDAI} &7B (Vicuna-v1.5)&1 fps&10.8 &28.0 &28.0 &19.3 &25.1 &0.7 &0.0 &0.0 &0.2 &4.0 &2.7 &14.7 &7.1 \\
        Qwen2-VL~\cite{wang2024qwen2}&7B (Qwen2)&1 fps&33.9 & \underline{98.0} &76.0 &33.3 &69.1 &16.0 &12.7 &8.7 &12.4 &26.0 &9.3 &24.7 &20.0 \\
        LLaVA-OneVision~\cite{Li2024LLaVAOneVisionEV}&7B (Qwen2)&64 frm&51.8 &88.7 &87.3 & {\bf 55.3} &77.1 &70.0 &50.0 &37.3 &52.4 &41.3 &8.7 &27.3 &25.8 \\
        Video-XL~\cite{shu2025video}&7B (Qwen2)&1 fps&61.6 & \underline{98.0} &93.3 &48.7 & \underline{80.0} & \underline{89.3} & \underline{77.3} & {\bf 75.3} & \underline{80.6} &38.7 &7.3 &26.0 &24.0 \\
        LLaVA-Video~\cite{zhang2024video}&7B (Qwen2)&64 frm&62.5 &90.0 &88.7 & \underline{52.0} &76.9 &78.7 & \underline{77.3} &67.3 &74.4 & {\bf 54.7} &11.3 & {\bf 42.7} & {\bf 36.2} \\
        \rowcolor{magenta!20} {\bf QViC-MF (Ours)}&7B (Qwen2)&1 fps& \underline{63.0} & {\bf 99.3} & \underline{98.7} &39.3 &79.1 &88.7 & {\bf 84.7} &66.7 &80.0 &40.7 & {\bf 16.0} &32.7 &29.8 \\
        \rowcolor{magenta!20} {\bf QViC-MF (Ours)}&7B (Qwen2)&2 fps& {\bf 64.7} & {\bf 99.3} & {\bf 100.0} &42.0 & {\bf 80.4} & {\bf 91.3} & {\bf 84.7} & \underline{69.3} & {\bf 81.8} & \underline{45.3} & \underline{14.0} & \underline{36.7} & \underline{32.0} \\
      \midrule
      \multicolumn{4}{l}{\!\textit{Long (60-180 sec.)}}  && \\
      \rowcolor{gray!40} GPT-4o~\cite{Achiam2023GPT4TR} &-&1 fps&56.3 &100.0 &98.0 &84.0 &94.0 &73.5 &80.0 &26.5 &60.0 &22.3 &2.0 &20.4 &14.9 \\
      \rowcolor{gray!40} Gemini 1.5 Pro~\cite{Reid2024Gemini1U} &-&1 fps&65.1 &100.0 &94.0 &68.0 &87.3 &90.0 &96.0 &34.0 &73.3 &56.0 &10.0 &38.0 &34.7 \\
        LLaMA-VID~\cite{Li2023LLaMAVIDAI}&7B (Vicuna-v1.5)&1 fps&6.4 &14.0 &16.0 &14.0 &14.7 &0.0 &0.0 &2.0 &0.7 &4.0 &0.0 &8.0 &4.0 \\
        Qwen2-VL~\cite{wang2024qwen2}&7B (Qwen2)&1 fps&33.6 & \underline{96.0} &80.0 &26.0 &67.3 &24.0 &12.0 &8.0 &14.7 &34.0 &6.0 &16.0 &18.7 \\
        LLaVA-OneVision~\cite{Li2024LLaVAOneVisionEV}&7B (Qwen2)&64 frm&36.0 &72.0 &62.0 & {\bf 40.0} &58.0 &42.0 &30.0 &26.0 &32.7 &24.0 & \underline{10.0} &18.0 &17.3 \\
        LLaVA-Video~\cite{zhang2024video}&7B (Qwen2)&64 frm&40.4 &70.0 &66.0 & \underline{38.0} &58.0 &42.0 &42.0 &32.0 &38.7 &36.0 & {\bf 14.0} &24.0 &24.7 \\
        \rowcolor{magenta!20} {\bf QViC-MF (Ours)}&7B (Qwen2)&1 fps& \underline{56.9} & {\bf 100.0} & \underline{96.0} &34.0 & \underline{76.7} & \underline{82.0} & \underline{70.0} & \underline{48.0} & \underline{66.7} & \underline{44.0} & \underline{10.0} & {\bf 28.0} & {\bf 27.3} \\
        \rowcolor{magenta!20} {\bf QViC-MF (Ours)}&7B (Qwen2)&2 fps& {\bf 58.7} & {\bf 100.0} & {\bf 100.0} & \underline{38.0} & {\bf 79.3} & {\bf 86.0} & {\bf 72.0} & {\bf 50.0} & {\bf 69.3} & {\bf 46.0} &8.0 & {\bf 28.0} & {\bf 27.3} \\
      \bottomrule
  \end{tabular}
  \caption{
    Detailed comparison of the proposed QViC-MF framework with SoTA methods on the VNBench.
    The best and second best results are highlighted in bold and underlined, respectively.
  }
  \label{tab:vnbench}
  \end{table*}

\section{Detailed Results on Benchmarks}
\label{sec:benchmark_details}

Tables~\ref{tab:mlvudev} to \ref{tab:vnbench} present the detailed benchmark results for 
MLVU-dev~\cite{Zhou_2025_CVPR}, LVBench~\cite{wang2025lvbench}, VideoMME~\cite{fu2025video}, 
and VNBench~\cite{zhao2025needle}. 
Across all tables, we use two notations to indicate the input sampling strategy of each method: 
(i) Uniform Sampling (``N~frm''), which evenly samples $N$ frames per video, and 
(ii) Frame Rate Sampling (``N~fps''), which samples videos at $N$ frames per second.

\paragraph{MLVU-dev.}
MLVU-dev~\cite{Zhou_2025_CVPR} evaluates diverse long-term video understanding tasks using videos 
ranging from 3 minutes to over 2 hours. 
Following the official protocol, we report the mean task accuracy (M-Avg). 
The benchmark consists of seven tasks: Topic Reasoning (TR), Anomaly Recognition (AR), 
Needle QA (NQA), Ego Reasoning (ER), Plot QA (PQA), Action Order (AO), and Action Count (AC). 
Compared with previous state-of-the-art (SoTA) methods, QViC-MF achieves a +1.7\% improvement in M-Avg. 
The gain is particularly pronounced in the AO task, which requires completing long-range temporal event sequences; 
QViC-MF improves AO accuracy by +11.2\%. This substantial gain indicates that the proposed 
memory-feedback mechanism effectively supports long-horizon temporal reasoning.

\paragraph{LVBench.}
LVBench~\cite{wang2025lvbench} serves as a long-term video understanding benchmark with videos 
up to roughly two hours in duration. 
Following the official evaluation protocol, we report the overall accuracy across all questions. 
The benchmark covers six tasks: Key Information Retrieval (KIR), 
Event Understanding (EU), Entity Recognition (ER), Temporal Grounding (TG), Reasoning (Rea), and Summarization (Sum). 
Compared with previous SoTA methods, QViC-MF achieves a substantial improvement of +8.3\% in overall accuracy.
The gains are particularly notable in four tasks requiring spatiotemporal focus and long-range event memory: 
+13.0\% in KIR, +8.2\% in EU, +10.1\% in ER, and +8.2\% in TG. 
These improvements suggest that QViC-MF’s memory-feedback mechanism effectively supports the retrieval 
and integration of key events over extended temporal spans.

\paragraph{VideoMME.}
VideoMME~\cite{fu2025video} covers multi-domain video comprehension with durations ranging from 
short clips to hour-long videos. 
We follow the video-only evaluation setting without subtitles and report the average accuracy across all samples. 
QViC-MF outperforms previous SoTA methods across all duration ranges.
The improvement is most pronounced for the Long range, where QViC-MF achieves a +3.7\% gain, 
demonstrating its advantage in understanding long-duration videos and 
maintaining consistent performance as the temporal horizon increases.

\paragraph{VNBench.}
VNBench~\cite{zhao2025needle} targets the challenging Needle-in-a-Haystack (NIAH) scenario and 
adopts a 4-try circular evaluation protocol, where a prediction is considered correct only if all 
four trials are answered accurately. 
We report results for both the full duration range ALL (10--180 sec) and the long-duration subset Long (60--180 sec), 
covering Retrieval, Ordering, and Counting tasks as well as their overall averages. 
QViC-MF achieves improvements of +2.2\% in the ALL overall score and 
a substantial +18.3\% in the Long overall score compared to prior SoTA methods. 
The gains in the long-duration setting are particularly large: +21.3\% in Retrieval and +30.6\% in Ordering. 
These results highlight the effectiveness of QViC-MF’s spatiotemporally selective compression 
and memory-feedback mechanisms in the challenging long-horizon NIAH scenario, 
where identifying sparse, question-relevant events is crucial.


\section{Further Ablation Studies}
\label{sec:ablation_studies}
The ablations in this section focus on hyper-parameters that are intrinsic to the core mechanisms of QViC-MF, 
namely the memory-feedback process and the question-guided visual compression module. 
These factors directly affect how the model retrieves long-range evidence and preserves question-relevant information, 
and are therefore essential for understanding the behavior and effectiveness of QViC-MF.
Tables~\ref{tab:feedback_ablation} to \ref{tab:memory_ablation} 
present the detailed ablation studies conducted in our work.
All experiments vary one specific parameter in each ablation 
while keeping all other settings identical to the inference configuration described in Table~\ref{tab:hyperparameter_inference}.
All results are reported using 2~fps input sampling.

\begin{table}[!t]
  \small
  \centering
  \begin{tabular}{l|ccc}
      \toprule
         & {\bf MLVU} & \multicolumn{2}{c}{\bf VNBench} \\
        {\bf $K, K_\mathrm{r}$}& test & ALL & Long \\
      \midrule
        64, 0&48.7 &62.4 &54.7 \\
        48, 16& \underline{58.3} &63.8 & {\bf 58.9} \\
        \cellcolor{magenta!20}32, 32& {\bf 59.4} & {\bf 64.7} & \underline{58.7} \\
        16, 48&58.0 & \underline{63.9} &57.1 \\
      \bottomrule
  \end{tabular}
  \caption{
    Ablation study on the balance between the number of current clip frames $K$ and recall frames $K_{\mathrm{r}}$.
    All other inference hyper-parameters follow the settings in Table~\ref{tab:hyperparameter_inference}.
    The magenta-highlighted configuration corresponds to the default setting used in Table~\ref{tab:hyperparameter_inference}.  
    The best and second-best results are highlighted in bold and underlined, respectively.
  }
  \label{tab:feedback_ablation}
\end{table}

\paragraph{Number of recall frames.}
Table~\ref{tab:feedback_ablation} presents the ablation study on the balance between 
the number of current clip frames $K$ and recall frames $K_{\mathrm{r}}$.
As noted in Section~\ref{sec:training_details}, the visual compressor operates with a 64-frame input,
and we divided this budget between $K$ and $K_{\mathrm{r}}$.
We observe that $K_{\mathrm{r}} = 0$ (no feedback) results in poor performance, 
especially for long videos, whereas introducing recall frames consistently improves accuracy.  
However, increasing $K_{\mathrm{r}}$ also decreases throughput: as $K$ becomes smaller, each 
64-frame window covers a shorter portion of the video, increasing the number of windows 
required to process the entire sequence.  
Balancing these two factors, the configuration $(K, K_{\mathrm{r}}) = (32,32)$ provides the 
best trade-off between accuracy and computational efficiency.

\begin{table}[!t]
  \small
  \centering
  \begin{tabular}{l|ccc}
      \toprule
        & {\bf MLVU} & \multicolumn{2}{c}{\bf VNBench} \\
        {\bf $K_\mathrm{h}$} & test & ALL & Long \\
      \midrule
        1&57.1 & {\bf 64.7} &57.8 \\
        \cellcolor{magenta!20}5 & {\bf 59.4} & {\bf 64.7} & {\bf 58.7} \\
        14&56.9 & {\bf 64.7} & \underline{58.2} \\
        28 (all heads)& \underline{57.6} & \underline{64.6} & 58.0 \\
      \bottomrule
\end{tabular}
  \caption{
    Ablation study on the number of heads $K_\mathrm{h}$ used in relevance score computation. 
    Setting $K_\mathrm{h} = 28$ corresponds to using all attention heads. 
    The magenta-highlighted configuration denotes the default setting used in 
    Table~\ref{tab:hyperparameter_inference}. 
    The best and second-best results are highlighted in bold and underlined, respectively.
  }
  \label{tab:headnumber_ablation}
\end{table}

\paragraph{Hyper-parameters for relevance score computation.}
Table~\ref{tab:headnumber_ablation} summarizes the ablation results 
for the number of attention heads $K_\mathrm{h}$ used in computing the relevance score $r_{n,i}$ in Eq.~(4), 
and Table~\ref{tab:layernumber_ablation} presents the corresponding ablation for the layer range $[L_1, L_2]$.  
The results show that using $K_\mathrm{h} = 5$ and layers $[L_1, L_2] = [17, 20]$ yields the best overall performance.
The difference between MLVU and VNBench can be attributed to how frequently the context memory is updated during inference.  
MLVU contains many long videos whose total number of frames often exceeds the context memory capacity $L = 256$.  
In such cases, the model repeatedly prunes and updates memory entries making performance more sensitive to 
the quality of the relevance-score estimation.  
This explains why the choice of $K_\mathrm{h}$ and $[L_1, L_2]$ has a more noticeable impact on MLVU.
In contrast, most videos in VNBench fall within the 256-frame limit, so entry pruning rarely occurs.  
As a result, the relevance score plays a smaller role during inference, and the performance 
differences across configurations become correspondingly smaller.

\begin{table}[!t]
  \small
  \centering
  \begin{tabular}{l|ccc}
      \toprule
        & {\bf MLVU} & \multicolumn{2}{c}{\bf VNBench} \\
        {\bf $[L_1, L_2]$} & test & ALL & Long \\
      \midrule
        $[1, 28]$ (all layers) & 55.8 & {\bf 64.9} & {\bf 58.9} \\
        \cellcolor{magenta!20}$[17, 20]$ & {\bf 59.4} & 64.7 & 58.7\\
      \bottomrule
  \end{tabular}
  \caption{
    Ablation study on the layer range $[L_1, L_2]$ used in relevance score computation.
    Setting $[L_1, L_2] = [1, 28]$ corresponds to using all layers. 
    The magenta-highlighted configuration denotes the default setting used in 
    Table~\ref{tab:hyperparameter_inference}.
    The best results are highlighted in bold.
  }
  \label{tab:layernumber_ablation}
\end{table}

\begin{table}[!t]
  \small
  \centering
  \begin{tabular}{l|ccc}
      \toprule
         & {\bf MLVU} & \multicolumn{2}{c}{\bf VNBench} \\
         {\bf $C$} & test & ALL & Long \\
      \midrule
        1&54.5&56.7&51.6\\
        4&54.3& {\bf 65.3} & {\bf 60.2} \\
        \cellcolor{magenta!20}16 & {\bf 59.4} &64.7&58.7\\
      \bottomrule
  \end{tabular}
  \caption{
    Ablation study on the number of context tokens per frame $C$.
    The magenta-highlighted configuration denotes the default setting used in 
    Table~\ref{tab:hyperparameter_inference}. 
    The best results are highlighted in bold.
  }
  \label{tab:contexttoken_ablation}
\end{table}

\begin{table}[!t]
  \small
  \centering
  \begin{tabular}{l|ccccc}
      \toprule
        {\bf $L$} & {\bf M-Avg} &  {\bf NQA} & {\bf ER} & {\bf AO} & {\bf TQA}  \\
      \midrule
        32&56.1&68.3& \underline{73.6} &48.6&48.8\\
        64& \underline{58.1} & {\bf 76.7} & {\bf 79.3} & \underline{54.3} &44.2\\
        128&57.7& \underline{73.3} & {\bf 79.3} & \underline{54.3} & \underline{46.5} \\
        \cellcolor{magenta!20}256 & {\bf 59.4} &71.7&71.7& {\bf 61.4} & {\bf 55.8} \\
      \bottomrule
  \end{tabular}
  \caption{
    Ablation study on the context memory capacity $L$. 
    We report results on the Needle QA (NQA), Ego Reasoning (ER), Action Order (AO), and Tutorial QA (TQA) tasks of MLVU-test, 
    together with their mean average (M-Avg). 
    The magenta-highlighted configuration denotes the default setting used in 
    Table~\ref{tab:hyperparameter_inference}. 
    The best and second-best results are highlighted in bold and underlined, respectively.
  }
  \label{tab:memory_ablation}
\end{table}

\paragraph{Number of context tokens.}
Table~\ref{tab:contexttoken_ablation} shows the effect of varying the number of context tokens per frame $C$.
For MLVU, performance drops sharply once $C$ is reduced from 16 to 4 or 1.
This trend suggests that the video content and tasks in MLVU, 
which often involve complex actions and temporal dependencies, 
require sufficiently detailed per-frame representations. 
With a larger value of $C$ (e.g., $C = 16$),
the model has enough capacity to capture the necessary visual information within each frame, 
whereas reducing $C$ limits this capacity and leads to a loss of important cues.
In contrast, VNBench remains relatively stable when $C$ is reduced from 16 to 4, 
with only a substantial performance degradation occurring at $C = 1$.
This robustness reflects the nature of VNBench, where the relevant visual evidence is typically simpler 
(e.g., static scenes or short text-like cues), 
making the tasks less demanding in terms of per-frame representation quality.
These results indicate that, for general long-video understanding scenarios as in MLVU, 
a moderate number of context tokens (e.g., $C = 16$) is preferable, 
providing sufficient capacity to encode frame-level information while maintaining stable downstream performance.

\paragraph{Context memory capacity.}
We analyze the effect of the context memory capacity $L$ 
on the Needle QA (NQA), Ego Reasoning (ER), Action Order (AO), and Tutorial QA (TQA) tasks of MLVU-test,
as shown in Table~\ref{tab:memory_ablation}. 
Overall, the best mean performance (M-Avg) is achieved at $L = 256$, while $L = 64$ and $L = 128$ yield comparable averages.
However, the optimal capacity depends on how the task distributes question-relevant information over time.
For NQA and ER, where the crucial evidence is concentrated in a relatively small number of frames
(e.g., a short needle action or a specific event/instance), 
smaller memory capacities ($L = 64$ or $L = 128$) tend to perform better, 
as they can retain the key frames while reducing the proportion of irrelevant distractor memories.
In contrast, AO and TQA involve multiple actions or instances spread over longer temporal spans,
and thus benefit from a larger memory capacity ($L = 256$), 
which reduces the risk of missing important frames even at the cost of storing more distractors.
These results suggest that the trade-off between preserving all question-relevant evidence 
and suppressing distractor memories is crucial when choosing the memory capacity.
As a direction for future work, mechanisms such as explicit forgetting or 
pruning of unused memories could help mitigate the impact of distractors 
when using large memory capacities.

\begin{figure*}[!t]
  \centering
  \includegraphics[width=1.0\linewidth]{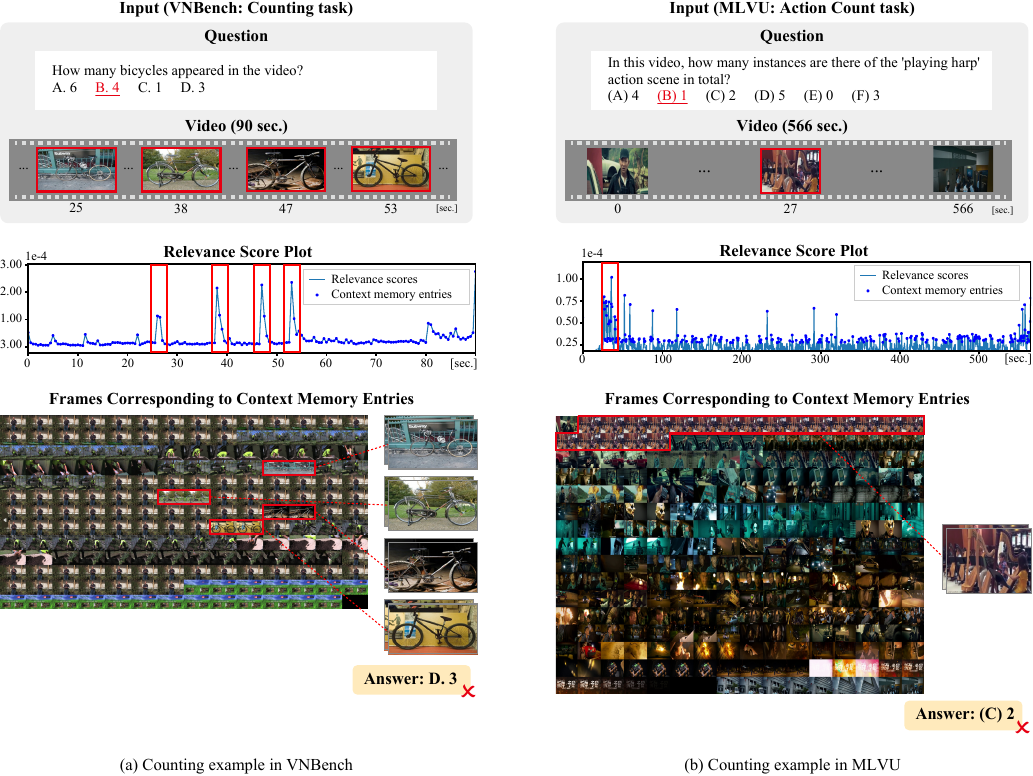}
  \caption{
    QViC-MF failure cases on counting tasks from VNBench and MLVU.
    For each example, the input question, relevance score plot, 
    and the frames corresponding to context memory entries are shown.
    (a)~VNBench example:
    the model correctly identifies and stores all four bicycle-containing segments in the context memory, 
    yet it fails to produce the correct count in its final answer.
    (b)~MLVU example:
    the model accurately detects and stores the single occurrence of the ``playing harp'' action, 
    but it likewise fails to output the correct count.
  }
  \label{fig:failure_cases}
\end{figure*}

\section{Computational Complexity}
\label{sec:complexity}

Given a video of length $T$, QViC-MF processes it sequentially in fixed-size clips.
At each step, the visual encoder and compressor operate on a bounded window
of $K_\mathrm{v} = K + K_\mathrm{r}$ frames, resulting in a bounded encoder token length
$N_{\text{enc}}$ independent of $T$.
Both the context memory and decoding operate on a fixed token budget
independent of $T$, since the memory capacity is fixed and the decoder is invoked once.
As a result, the total end-to-end computational complexity scales
linearly with the video length, i.e., $O(T)$.
Memory feedback may re-encode recalled frames, but the total number
of encoded frames scales as $(1 + K_\mathrm{r}/K)T$, which incurs only a
constant-factor overhead compared with one-way frame-wise methods.
This linear-in-$T$ scaling is also shared by recent memory-based
approaches such as Flash-VStream~\cite{Zhang_2025_ICCV}, although
architectural choices and constant factors differ.

\section{Inference Time and VRAM Usage}
\label{sec:inference_time_vram}

On a system equipped with an H200 GPU (141~GB VRAM),  
Intel(R) Xeon(R) Platinum 8558 CPU @4.00GHz, and 3.0~TiB RAM,  
QViC-MF can construct the context memory at 28 frames per second.  

As mentioned in the main paper, 
QViC-MF uses the same base LLM (Qwen2-7B~\cite{yang2024qwen2technicalreport}, 15~GiB)  
for both the visual compressor and the decoder,  
and applies lightweight tuning with LoRA~\cite{Hu2021LoRALA} (0.7~GiB per LLM).  
By sharing LLM parameters, the total VRAM usage of QViC-MF can be kept around 16.4~GiB,  
comparable to typical LMMs such as LLaVA-Video-7B-Qwen2~\cite{zhang2024video}.  
In the current implementation, however, 
the visual compressor and decoder are instantiated as separate models,  
resulting in a total VRAM usage of 30.7~GiB.
This can be reduced through the shared-parameter implementation described above.
Each context embedding occupies less than 1~MiB,  
so storing several hundred embeddings in the context memory increases 
VRAM usage by less than 1~GiB.  
QViC-MF compresses input videos by sequentially adding context embeddings  
to a fixed-length context memory.  
This design ensures that VRAM usage scales with the memory length $L$,
enabling stable resource usage even as video length increases.

Overall, QViC-MF achieves both efficient inference speed and low VRAM consumption 
for practical applications in long-term video understanding, thanks to its memory-efficient design.

\section{Reliability of Relevance-based Memory Retrieval}
\label{sec:memory_stability}

The relevance scores used in QViC-MF are derived from internal attention statistics
and serve as lightweight routing signals for memory retention and retrieval.
While such signals may be imperfect, the design of QViC-MF prevents error amplification.
The context memory has a fixed capacity with top-$K_\mathrm{r}$ retrieval,
and recalled frames are re-encoded together with the current clip,
updating only their corresponding memory slots.
These mechanisms prevent self-reinforcing feedback loops.

To empirically evaluate the reliability of relevance-based retrieval,
we measure the needle hit-rate on VNBench Long using annotated needle timestamps.
To avoid trivial hits from storing all frames,
we evaluate only videos longer than 60 seconds,
sample input frames at 2 fps,
and constrain the context memory to 64 frames.
We report the sample-level needle hit-rate,
where a sample is counted as correct only if all needles in the video
are retained in the context memory.
Under this constrained setting,
relevance-based feedback achieves a 99.8\% hit-rate (1796/1800 samples),
whereas uniform sampling with the same 64-frame memory budget
achieves only 38.7\% (696/1800 samples),
indicating that retrieval failures are rare in practice.

%
%
%
%

\section{Discussion and Limitations}
\label{sec:limitation}

\paragraph{Effectiveness of clip-level memory feedback.}
Clip-level memory feedback is not uniformly beneficial across all question types. 
Our evaluation includes tasks that require re-localizing sparse events 
and reasoning about their temporal order (e.g., VNBench and MLVU Action Order), 
as well as tasks that involve more global or descriptive understanding 
(e.g., MLVU Topic Reasoning and LVBench Summarization).

We observe that the proposed feedback mechanism yields the largest gains 
when questions require retrieving sparse, question-relevant events from long videos.
In such cases, single-pass compression may fail to preserve the complete event, 
whereas memory feedback enables the model to revisit and refine relevant visual evidence across clips.
In contrast, for global or summary-style questions that rely on
distributed visual cues across the video, the benefits of feedback are
smaller because answering these questions requires retaining
broad contextual information rather than a small set of key events.
Future work may explore hierarchical memory architectures that
simultaneously maintain local event-level representations and
global video-level summaries, enabling the model to better support
both sparse event retrieval and holistic video understanding.

\paragraph{Counting tasks.}
Despite the overall performance improvements achieved by QViC-MF, 
its effectiveness on counting tasks remains limited. 
As shown in Tables~\ref{tab:mlvudev} and \ref{tab:vnbench},
QViC-MF does not exhibit noticeable gains on tasks that require counting the number of specific visual events or object occurrences,
such as the Action Count (AC) task in MLVU and the Counting task in VNBench. 
We note that this limitation is not unique to QViC-MF; 
counting remains a challenging task even for strong proprietary models 
such as GPT-4o~\cite{Achiam2023GPT4TR} and Gemini~1.5~Pro~\cite{Reid2024Gemini1U},
as well as for existing open-source LMMs.

We present representative failure cases for the counting tasks in VNBench and MLVU in Figure~\ref{fig:failure_cases}. 
In these examples, QViC-MF successfully identifies and stores the frames corresponding to the events to be counted, 
yet the decoder LLM still fails to output the correct count. 
This suggests that the bottleneck lies not in the visual retrieval or memory mechanism 
but in the final reasoning stage performed by the language model.
As a direction for future work, expanding training data that explicitly targets counting, 
or incorporating counting-specific mechanisms such as explicit video segmentation or instance tracking, 
may further enhance the counting ability of LMMs broadly.

\end{document}